\documentclass[a4paper]{cas-dc}

\usepackage{tikz}
\usetikzlibrary{backgrounds}
\usepackage{gensymb}
\usepackage{svg}
\usepackage{graphicx}
\DeclareGraphicsExtensions{.png,.pdf}
\usepackage{float}
\usepackage{subcaption}
\usepackage{amsmath}
\usepackage{adjustbox}
\usepackage{algorithm}
\usepackage{algpseudocode}
\usetikzlibrary{shapes.geometric, arrows,calc,patterns,angles,quotes}
\usepackage{xcolor}
\usetikzlibrary{positioning}
\usepackage[authoryear,longnamesfirst]{natbib}
\usepackage{nomencl}
\makenomenclature
\usepackage{multirow}
\usepackage{soul}
\usepackage{lipsum}

\def\tsc#1{\csdef{#1}{\textsc{\lowercase{#1}}\xspace}}
\tsc{WGM}
\tsc{QE}

\newcommand{\Secref}[1]{\hyperref[#1]{Section~\ref*{#1}}}

\newcommand{\rev}[1]{\color{black}#1\color{black}}

\begin{document}
\let\WriteBookmarks\relax
\def\floatpagepagefraction{1}
\def\textpagefraction{.001}

\shorttitle{Collision Avoidance using APFs}    

\shortauthors{Aditya, Abhilash}  

\title [mode = title]{Collision Avoidance for Autonomous Surface Vessels using Novel Artificial Potential Fields}

\author[aff1]{Aditya Kailas Jadhav}[orcid=0000-0001-6806-1537]
\ead{aditya.jadhav@smail.iitm.ac.in}



\ead[url]{https://adityajadhav99.github.io}


\affiliation[aff1]{organization={Marine Autonomous Vehicles Laboratory, Department of Ocean Engineering},
            addressline={Indian Institute of Technology (IIT) Madras}, 
            city={Chennai},
            country={India},
            pin={ - 600036}}

\affiliation[aff2]{organization={Center for Maritime Experiments to Maritime Experience (ME2ME)},
            addressline={Indian Institute of Technology (IIT) Madras}, 
            city={Chennai},
            country={India},
            pin={ - 600036}}






\author[aff1, aff2]{Anantha Raj Pandi}
\ead{oe23d007@smail.iitm.ac.in}

\author[aff1, aff2]{Abhilash Somayajula}[orcid=0000-0002-5654-4627]
\cormark[1]
\ead{abhilash@iitm.ac.in}
\ead[url]{https://www.doe.iitm.ac.in/abhilash}

\cortext[co]{Corresponding author.}



\begin{abstract}
    As the demand for transportation through waterways continues to rise, the number of vessels plying the waters has correspondingly increased. This has resulted in a greater number of accidents and collisions between ships, some of which lead to significant loss of life and financial losses. Research has shown that human error is a major factor responsible for such incidents. The maritime industry is constantly exploring newer approaches to autonomy to mitigate this issue. This study presents the use of novel Artificial Potential Fields (APFs) to perform obstacle and collision avoidance in marine environments. This study highlights the advantage of harmonic functions over traditional functions in modeling potential fields. With a modification, the method is extended to effectively avoid dynamic obstacles while adhering to COLREGs. Improved performance is observed as compared to the traditional potential fields and also against the popular velocity obstacle approach. A comprehensive statistical analysis is also performed through Monte Carlo simulations in different congested environments that emulate real traffic conditions to demonstrate robustness of the approach.
\end{abstract}


\maketitle

\section{Introduction}
Owing to an increase in inter-continental trade, the demand for transportation through waterways has grown significantly in recent years. This increase in traffic in both inland waterways and the seas has resulted in a higher frequency of accidents and ship collisions that cause significant financial losses and even pose a risk to lives. Errors in human decision-making are a major cause of accidents in the maritime industry. \cite{baker2005accident} analyzed the maritime accident investigations and concluded that around 80\% of marine accidents are caused by the human element, especially the failure of situation awareness and situation assessment. Human fatigue was also identified as one of the major reason for the lack of situational awareness. One notable incident highlighting the importance of autonomy in the marine environment is the Ever Given disaster that occurred in the Suez Canal in March 2021. The Suez Canal, one of the world's busiest trade routes, was blocked for six days by the Ever Given, a 400-meter-long container ship. The vessel ran aground due to strong winds, wedging itself across the canal and blocking all traffic. Technical or human errors were suspected to have contributed to the incident. The obstruction occurred in a section of the canal, causing significant disruptions to trade between Europe, Asia, and the Middle East, which cost global trade between \$6 billion to \$10 billion according to \cite{bbc_article}.

The Ever Given incident emphasized the need for advanced autonomy and obstacle avoidance systems. It highlighted the economic impact and reliance on safe and efficient navigation in waterways. The autonomous navigation of a marine vessel is a difficult problem, considering the fact that marine vessels are under-actuated and are sluggish to respond to an applied control action. While navigating in the waterways, all vessels have to comply with Convention on the International Regulations for Preventing Collisions at Sea \citep{std1972colregs}.

As the developments toward building of completely autonomous ships accelerate, it is important to ensure that the algorithms governing their behavior comply with COLREGS. In general, this can be achieved through a local guidance mechanism that is activated when the static and dynamic obstacles are detected within a certain specified distance of the ship. Some of the popular approaches for collision avoidance include rule based methods \citep{benjamin2002multi}, hybrid methods that integrate global path planning methods with collision avoidance \citep{chiang2018colreg}, reactive methods \citep{xue2011automatic} and optimization based methods \citep{szlapczynski2011evolutionary}. The recent era of artificial intelligence has also seen collision avoidance being tackled through the reinforcement learning paradigm \citep{zhao2019colregs, meyer2020colreg, sawada2021automatic}.

Some of the early work on the rule based approaches for following COLREG rules came in the studies of \cite{benjamin2006method} and led to the development of the MOOS-IvP framework for collision avoidance. However, most these early studies using rule based approaches focused on simple ship encounters and were not easy to extend to multiple ship encounters. This led to the exploration into the hybrid methods that aimed to integrate collision avoidance as a part of the global path planning task. Some of the approaches included the informed-RRT* \citep{enevoldsen2021COLREGS}, fast marching algorithm \citep{liu2017fast}. These methods are a family of algorithms used in robotics and motion planning that generate a set of random samples in the configuration space to efficiently search for feasible paths to move a robot from a starting position to a goal position as explained in \cite{lavalle1998rapidly}. Although these methods are probabilistically complete, i.e., given enough time, they are guaranteed to find a solution to a problem if one exists but they do not guarantee optimality, and the paths generated are non-deterministic. The algorithms are highly generalizable in nature as they can be applied for planning applications in higher dimensions. However, the computational cost associated with building a space-filling grid is quite high and hence these approaches are not quite popular for local planning problems.

In order to keep the computational costs manageable, several researchers started looking towards reactive methods. One of the popular approaches in the reactive methods is the velocity obstacle method \citep{huang2018velocity}. This method has been improved further to incorporate uncertainties \citep{cho2020efficient}. Similarly, further improvements have also been made to relax the assumption of constant velocity of the target ship by active sharing of trajectory between the vessels \citep{kufoalor2018proactive, huang2019generalized}. These methods are also computationally cheaper than the hybrid methods such as COLREGS-RRT. However, the main drawback is that it still relies on knowing the trajectory of the target ship to some degree during the construction of the forbidden velocity space. In addition, it primarily relies on the kinematics of the vessel and does not take into account the sluggish dynamics of the vessels. 

Some of the recent attempts at solving the collision avoidance problem have focused on optimization methods to solve the local path planning approach. While some approaches rely on the model predictive control framework to evaluate a solution \citep{johansen2016ship} others have aimed to utilize evolutionary algorithms to solve the optimization problem \citep{hu2017colregs, lazarowska2015ship}. The main drawback of these methods is that these methods are computationally expensive and cannot always be evaluated in real time applications.

With the success of artificial intelligence-based methods in many related areas \citep{zinage2021deep, alamAIonWater, josenavigating, sanjeevcomparison, deraj2023deep}, there has been a significant rise in studies investigating similar approaches for obstacle and collision avoidance. \cite{zhao2019colregs} proposed Deep Reinforcement Learning based approach for multi-ship collision avoidance problem in congested sea areas. \cite{shen2019automatic} proposed a Deep Q-Learning based approach for automatic collision avoidance of multiple ships in restricted waters. Their algorithm also incorporated human experience and navigation rules for planning. Model scale experiments were also demonstrated to validate the simulations. Other studies by  \cite{chun2021deep} applied the Proximal Policy Optimization (PPO) algorithm to implement COLREGs. They compared the performance of the PPO algorithm with that of the A* algorithm and observed that the maximum Collision Risk (CR) was significantly reduced with the PPO algorithm. While the artificial intelligence based methods show good promise, they have some limitations that limit their field implementation. These methods learn on the basis of reward functions defined by a designer and are not always firmly rooted in the physics. Thus, guarantees cannot be provided that the approach will work in all scenarios. In addition, understanding what these models have learnt from the data (explainability of AI) is still an open area of research.

One of the less explored reactive methods for collision avoidance of marine crafts is artificial potential fields (APFs). While this approach has seen several applications in robotics \citep{khatib1986real, kim1992real, cetin2013establishing, panati2015autonomous}, very few studies have investigated the application of APFs to collision avoidance of ships. \cite{modifiedapflyu} introduced modified APFs for collision avoidance in a static and dynamic obstacle environment. They modified the original attractive and repulsive potential fields suggested by \cite{khatib1986real} to include dependence on the relative velocities between the Own-Ship (OS) and the Target Ship (TS) to consider dynamic obstacles. \cite{lazarowska2020discrete} utilized a discrete APF to obtain a collision-free trajectory for an own ship in near real-time and further optimized the trajectory using a Path Optimization Algorithm (POA). Another approach was introduced by \cite{wang2018colregs} that uses introduces potential around obstacles using Gaussian probability density functions. The study showed that it has potential to consider static as well as dynamic obstacles and can be made COLREGS compliant. 

One of the limitation of applying APF in global path planning is that in some cases the algorithm can be stuck in a local minima and thus the agent or vessel cannot reach the goal. However, they are an effective candidate for a local reactive planning. The traditional APF formulation assumes a quadratic attractive potential and an inverse square repulsive potential. Although this is quite suitable for finding a collision free path for a manipulator robot, it has limitations when applied to collision avoidance of mobile robots. Particularly the quadratic attractive potential gradients can dominate the repulsive gradients when the goal is located far away. While this might not be a problem for manipulators are the configuration domain is mostly bounded, it poses some issues as will be discussed in section \ref{sec:apfs}. 




Although other grid-based and sampling-based approaches exist for local obstacle avoidance problem, APFs offer the advantage of fast real-time implementation that is not easily achievable using the other approaches. Similarly, most study of APFs has stemmed from the control of manipulators, wheeled robots, and UAVs (Unmanned Aerial Vehicles) and very few studies have investigated APFs for marine vessels that are characterized by large inertial dynamics. This study presents the application of harmonic functions based APF planner for local obstacle avoidance for an ASV with a non-holonomic motion model. The significant advantage of the proposed method over other reactive methods is demonstrated. Additionally, a comprehensive statistical analysis based on certain performance metrics is performed to demonstrate the robustness of the approach.
 
\Secref{sec:dynamics} describes the non-holonomic MMG motion model for an Autonomous Surface Vessel. A local planner for obstacle avoidance assumes that a global planner has already generated a path in terms of waypoints. \Secref{sec:wp_tracking} explains the guidance and control algorithms for the waypoint tracking capability of the ASV. The proposed candidate potential fields based on harmonic functions are introduced and analyzed in \Secref{sec:apfs}. \Secref{sec:dynamic_obs} describes the application of the new proposed APF for dynamic obstacle avoidance while following COLREG rules. This section also compares the performance of APF method against other methods. \Secref{sec:statistical_analysis} documents the results of a comprehensive statistical analysis to demonstrate robustness of the approach. Finally, conclusions and directions for future research are provided in \Secref{sec:conclusion_futurework}. 
\section{Ship Dynamics}
\label{sec:dynamics}
This section describes the Maneuvering Modelling Group (MMG) model that is used to simulate the ASV dynamics.
For this study, the KRISO Container Ship (KCS) is chosen as the own ship (OS) for which the obstacle avoidance capabilities have been set up.

The simulation study is formulated using two coordinate systems: the earth-fixed global coordinate system (GCS) and the body coordinate system (BCS). While GCS is an inertial frame of reference, the BCS is a non-inertial frame that is fixed to the vessel and moves with it. The origin of the BCS is located at the intersection of the midship, centerline, and waterplane of the vessel, with the x-axis pointed towards the bow, the y-axis pointed towards the starboard side, and the z-axis pointed towards the keel (down) of the vessel. A general floating body can exhibit motions in all six degrees of freedom. However, as this study deals primarily with maneuvering motions of vessel, only the horizontal modes of motion (surge, sway and yaw) are modelled. 

\begin{figure}[htbp]
    \centering
    \includegraphics[width= \linewidth]{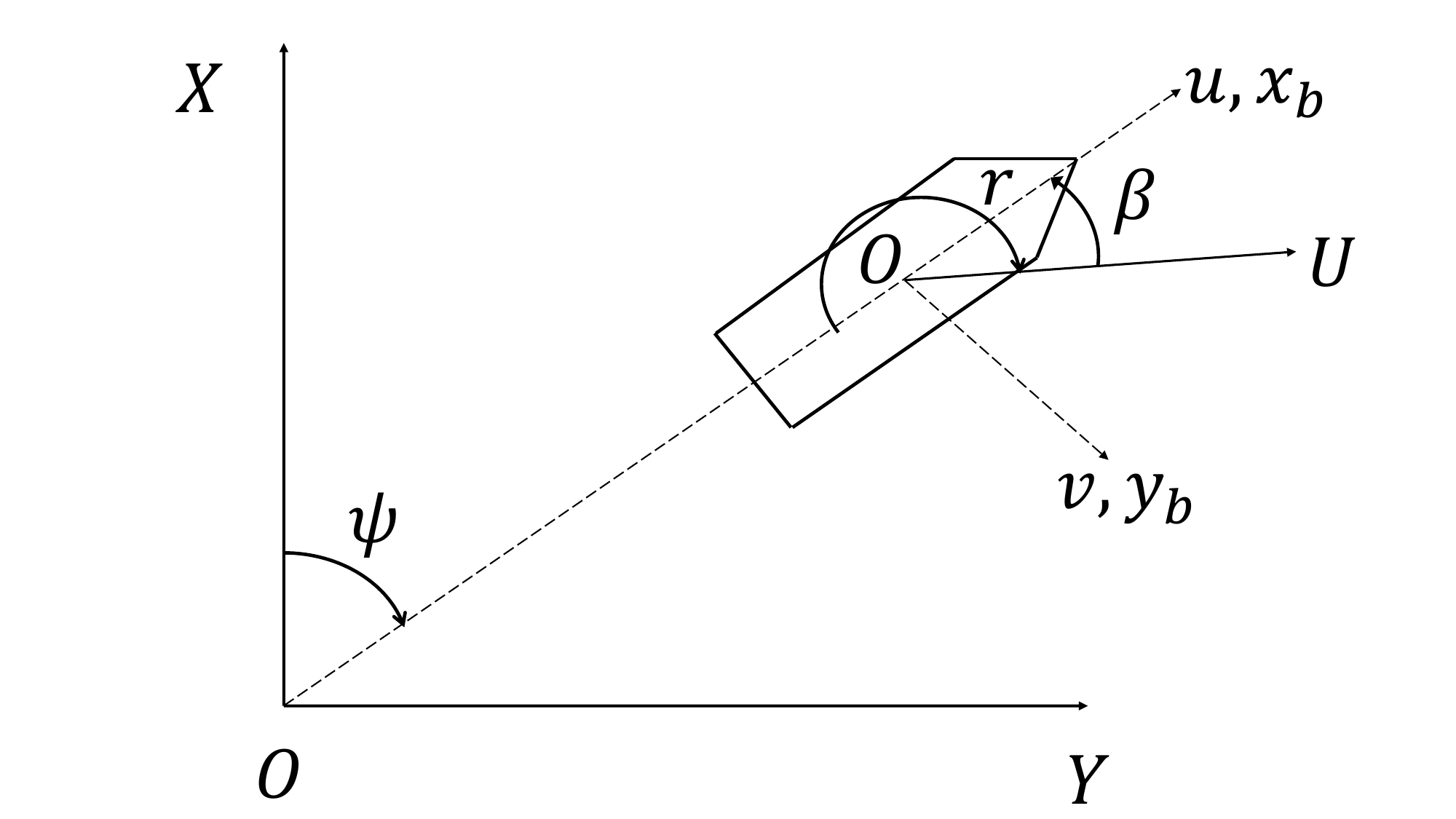}
    \caption{Coordinate Frames }
    \label{fig:gcs_bcs}
\end{figure}

The position and orientation of the vessel are denoted by $ \boldsymbol{\eta} = [x , y ,\psi]^T$ where $x$ and $y$ denote the position of the origin of BCS expressed in GCS and $\psi$ represents the heading angle of the ship with respect to the global x-axis. The velocity of the origin of the BCS, expressed in BCS frame is denoted by $\boldsymbol{\nu} = [u, v, r]^T$ where \textit{u}, \textit{v} and \textit{r} denote the surge velocity, sway velocity, and yaw rate. \autoref{fig:gcs_bcs} shows the GCS and the BCS coordinate frames. The kinematic relationship between these frames is given by \eqref{kinematics}.
\begin{equation}
    \boldsymbol{\dot \eta} = R(\psi) \boldsymbol{\nu}
    \label{kinematics}
\end{equation}

where $R(\psi)$ is the rotation matrix that transforms a vector from BCS to GCS. The rotation matrix depends on the heading angle of the vessel $\psi$ and is expressed as seen in \eqref{eq:rotation_matrix}.
\begin{equation}
R(\psi) = \begin{bmatrix}
            \cos(\psi) & -\sin(\psi) & 0 \\
            \sin(\psi) & \cos(\psi)  & 0 \\
            0           &         0  & 1
           \end{bmatrix}
    \label{eq:rotation_matrix}
\end{equation}

There are two popular approaches for modelling the maneuvering dynamics of a vessel - Abkowitz model \citep{abkowitz1980measurement} and MMG model \citep{yoshimura2012hydrodynamic}. In this study the vessel dynamics are modeled using the MMG approach. The Krisco Container Ship (KCS) is chosen for simulation as it is a benchmark hullform and its hydrodynamic parameters are readily available in the literature. The particulars of the full-scale Panamax Class KCS vessel are provided in \autoref{tab:KCSpara}.
\begin{table}[htbp]
    \renewcommand{\arraystretch}{1.2}
\centering
\caption{Full-Scale KCS ship parameters}
\begin{tabular*}{\tblwidth}{@{} LL@{} }
    \toprule
    \textbf{Ship Parameter} & \textbf{Value}\\
    \midrule
        Length between perpendiculars ($L$)& $230\,m$\\
        Length overall ($L_{OA}$)& $232.5\,m$\\
        Depth ($d$)& $19\,m$ \\
        Beam ($B$)& $32.2\,m$\\
        Draft ($d_{em}$)& $10.8\,m $ \\
        Displacement &$52030\,m^3$  \\
        LCG ($x_G$) &$-3.408\,m$\\
        Yaw radius of gyration & $57.5\,m$\\
        Design speed ($U$) & $12.347\,m/s$\\
    \bottomrule
   \end{tabular*}
\label{tab:KCSpara}
\end{table}

The dynamic equations of motion in surge, sway and yaw are given by \eqref{dynamics} where $m$ and $I_{zz}$ represent the mass and mass moment of inertia of the vessel along the z-axis of the BCS respectively, $m_x, m_y$, and $J_{zz}$ represent the added mass and added mass moment of inertia, $x_G$ denotes the longitudinal position of the center of gravity of vessel with respect to the BCS origin. The non-dimensional mass and added mass of the KCS vessel used in this study are reported in \autoref{tab:mass}.
\begin{table}[htbp]
    \renewcommand{\arraystretch}{1.2}
    \caption{Ship parameters}
    \begin{tabular*}{\tblwidth}{@{} LL@{} }
        \toprule
        \textbf{Parameter}&\textbf{Non-dimensional value}\\
        \midrule
        Surge added mass ($m_{x}$)&0.006269\\
        Sway added mass ($m_{y}$)&0.155164\\
        Yaw added mass moment ($J_{zz}$)&0.009268\\
        Yaw mass moment of inertia ($I_{zz}$)&0.011432\\
        Mass of the vessel ($m$)&0.182280\\
        \bottomrule
       \end{tabular*}
    \label{tab:mass}
\end{table}
\begin{eqnarray}
\begin{aligned}
    (m + m_x)\dot{u} - mvr - mx_G r^2 = X \\
    (m + m_y)\dot{v} + mx_G\dot{r} + mur = Y \\
    (I_{zz} + J_{zz})\dot{r} + mx_G\dot{v} + mx_G ur = N
    \label{dynamics}
\end{aligned}
\end{eqnarray}

$X$, $Y$, and $N$ are the external forces and moment acting on the vessel and can further be decomposed into components relating to the propeller, the rudder, and hull hydrodynamic forces and is shown in \eqref{tau}. In \eqref{tau} the subscripts $H$, $R$ and $P$ represent the hull, rudder, and propeller effects respectively. 

\begin{align}
\label{tau}
\begin{split}
      X =& \ X_{H}+X_{R}+X_{P}\\
     Y =& \ Y_{H}+Y_{R}\\
     N =& \ N_{H}+N_{R}
\end{split}
\end{align}

The propeller force is modelled by curve fitting the data from propeller open water characteristics. The rudder forces are calculated using aerofoil theory. The hull hydrodynamic forces are calculated through a sum of truncated Taylor series expansion whose coefficients are calculated from various standard experiments such as straight line runs in a towing tank or a planar motion mechanism (PMM) test. Although in general these hull hydrodynamic coefficients are determined by experiments, they can also be identified by applying system identification techniques to data collected from free-running models \citep{vijayidentification,deogaonkardatadriven}. Further details of the modeling of the propeller, rudder, and hull hydrodynamic forces can be found in the works of \cite{yoshimura2012hydrodynamic} or  \cite{deraj2023deep} and are omitted here for brevity.
 
 The surge and sway equations are non-dimensionalized by dividing both sides by $\frac{\rho}{2}U^2Ld_{em}$ where $\rho$ is the density of seawater, $L$ is the length of the vessel, $U$ is the design speed of the vessel and $d_{em}$ is the draft of the vessel. Similarly, the yaw equation of motion is non-dimensionalized by dividing both sides by $\frac{\rho}{2}U^2L^2d_{em}$. This system of non-dimensionalization is known as the prime-II system of normalization, as described in \cite{fossen2011handbook}. The non-dimensionalized equations are given by \eqref{eq:mmg_nd}
\begin{equation}
    \begin{split}
        &(m'+m_{x}')\dot{u}' - m' v' r' -  m' x_G' {r'}^2=X'\\
        &(m'+m_{y}')\dot{v}' + m' x_{G}'\dot{r}' + m' u' r'=Y'\\
        &(I_{zz}'+J_{zz}')\dot{r}' + m' x_{G}' \dot{v}' + m' x_{G}' u' r'=N'\\
    \end{split}
    \label{eq:mmg_nd}
\end{equation}

where $(.)'$ denotes a non-dimensionalized quantity. For keeping the notation uncluttered, prime notation is dropped from here on. However, it is understood that the quantities described are non-dimensional unless specified otherwise.


\section{Waypoint Tracking}
\label{sec:wp_tracking}
This section describes the guidance and control algorithms used in this study to achieve waypoint tracking for the underactuated dynamics as defined in \autoref{sec:dynamics}. Most marine vessel paths can be discretized into a set of waypoints and straight lines joining these waypoints. Waypoints are a specified set of coordinates, either Geographical (Latitude, Longitude) or Cartesian (X, Y), that are to be tracked by the vessel. In general these waypoints are given out by a global planner that has a static map of the environment to avoid the boundaries or static obstacles. However, sometimes there may be static obstacles such as stranded vessels that may not be present on the static map used by the planner.

The development of the global planner is not considered within the scope of this study. In this study, we presume that the waypoints to be tracked are already known from a global path planner and the vessel is tasked to track these waypoints. It is also pertinent to mention that this study is not aimed towards navigation in narrow channels that usually require evaluation of grounding risk through additional information that is not considered in this study. The aim of this study is to investigate static and dynamic collision avoidance of a vessel, where it is presumed that sufficient space is available for circumventing the static and dynamic obstacles.

Waypoint tracking of marine vessels is a classical field which has been extensively studied in the past decades. Most approaches separate the problem into two sub-problems - guidance and control. Guidance algorithm uses the waypoint location and current location of the vessel to estimate a desired heading angle for the vessel. The control algorithm applies the rudder to achieve the desired heading. In general there is also a navigation module that estimates the state of the vessel based on a mathematical model and sensor measurements using an filter (Extended Kalman Filter, Unscented Kalman Filter,  particle filter etc.). However, in this simulated study, a perfect state estimation is assumed and only the guidance and control algorithms are relevant.

Two of the popular waypoint guidance mechanisms for marine crafts are the Line of Sight (LOS) guidance \citep{caharija2016integral} and vector field guidance \citep{woo2019deep}. The vector field guidance is based on principles of sliding mode control and can lead to some chattering. In this study an integral-LOS guidance scheme is used that introduces a modification to the classical LOS guidance to include an integrator in a manner such that Lyapunov stability is maintained. This guidance approach is detailed in the following section.

Similarly, there are several different controller options to choose from for a marine craft such as - PID, linear quadratic regulator (LQR) \citep{morawski1998ship}, sliding mode control (SMC) \citep{mohan2023analyzing}, backstepping control \citep{mohan2023comparison}, model predictive control \citep{oh2010path}. In this study a simple PD controller is chosen to achieve the desired heading by driving the heading error to zero.

\subsection{ILOS Guidance}

Line of sight (LOS) guidance scheme is a conventional guidance scheme widely used for autonomous surface vessels. The key idea behind the LOS guidance scheme is to orient the heading of the vessel towards a point that is some fixed distance ahead along the line joining two successive waypoints. 
However, this scheme can exhibit instability if the look-ahead distance is poorly chosen and lacks effective disturbance compensation. Although such steady-state errors can be rejected through the introduction of an integral term in the LOS guidance scheme, the resulting guidance scheme is vulnerable to windup issues and also instability when the gains are not chosen appropriately. Thus, to overcome this limitation of typical integral terms, the ILOS scheme was proposed.  The ILOS scheme incorporates an integrator term in the guidance but its dynamics are chosen in such a way that the path along the track joining the waypoints is rendered globally asymptotically stable \citep{fossen2011handbook}.
The theoretical details of guidance implementation follow the works of \cite{caharija2016integral} and are reported here for completeness.

\begin{figure*}[h!]
  \centering
  \includegraphics[width= 0.8\textwidth]{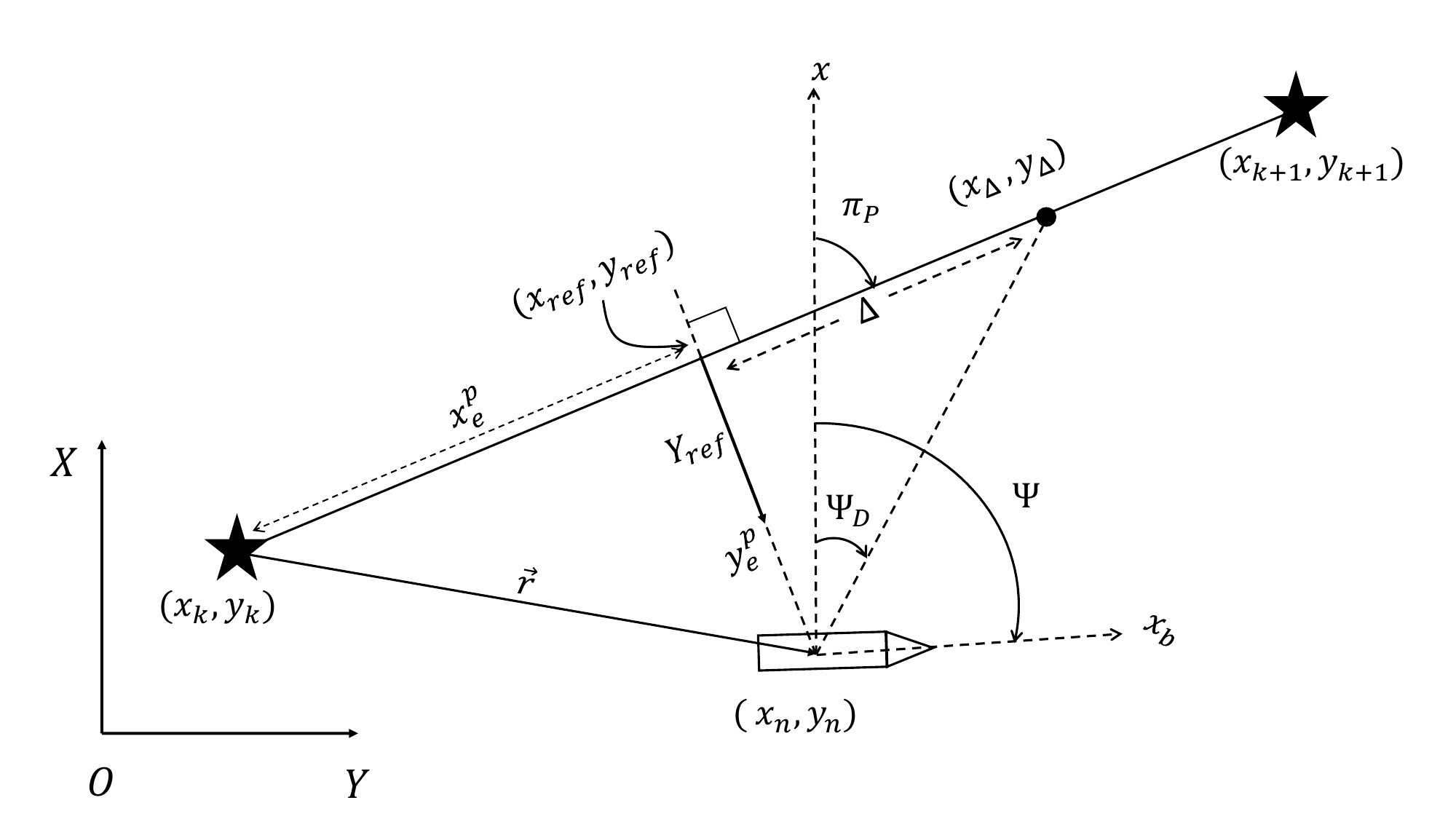}
    \caption{ILOS Guidance}
    \label{fig:I-LoS_guidance}
\end{figure*}

\autoref{fig:I-LoS_guidance} gives a pictorial representation of the scenario and the constituent parameters of this guidance law. 
Points $\left(x_k,y_k\right)$ and $\left(x_{k+1},y_{k+1}\right)$ denote the  coordinates of successive waypoints, where the $\left(k\right)^{\text{th}}$ waypoint has been tracked and the vessel is currently tracking the $\left(k+1\right)^{\text{th}}$ waypoint. The path tangential angle $\pi_p$, which denotes the angle between the line joining these waypoints and the X axis of the GCS, is given by \eqref{eq:pi_p}.

\begin{equation}    
    \pi_p =  \tan^{-1} \left(\frac{y_{k+1} - y_k}{x_{k+1} - x_k}\right) \label{eq:pi_p}
\end{equation}

The ship coordinates in GCS are expressed by $\left(x_n, y_n\right)$, and its heading with respect to the global X-axis is denoted by $\psi$. A reference point $\left(x_{ref}, y_{ref}\right)$ is defined at the foot of the perpendicular from the vessel's current position to the line joining the waypoints.  A frame of reference $\mathcal{X}_{ref}$ is defined with its origin at the reference point, the X-axis along the line joining  $\left(k\right)^{\text{th}}$ and $\left(k+1\right)^{\text{th}}$ waypoints and Y-axis as shown in \autoref{fig:I-LoS_guidance}. $\Delta$ denotes the chosen `look-ahead distance' parameter, which marks the point $\left(x_{\Delta},y_{\Delta}\right)$ ahead of the reference point, along the path to be tracked. In this study, $\Delta$ has been chosen to be twice the ship length. The cross-track error $y^p_e$ is the distance of the vessel from the reference point and quantifies the deviation of the vessel from the desired straight line path joining $\left(k\right)^{\text{th}}$ and $\left(k+1\right)^{\text{th}}$ waypoint. The distance of the reference point from the $\left(k\right)^{\text{th}}$ waypoint is called the along-track distance $x^p_e$. To compute the cross-track error $y^p_e$ and along-track distance $x^p_e$, consider the free vector $\vec{r}$ in \autoref{fig:I-LoS_guidance}.
When expressed in GCS, it can be written as seen in \eqref{eq:r_in_gcs}
\begin{equation}
\boldsymbol{r}^{GCS} = \boldsymbol{X_n} - \boldsymbol{X_k}
\label{eq:r_in_gcs}
\end{equation}
where $\boldsymbol{X_n}= \begin{bmatrix}
            x_n, y_n
        \end{bmatrix}^T$ is the current location of the vessel and $\boldsymbol{X_k} = \begin{bmatrix}
            x_k, y_k
        \end{bmatrix}^T$ is the location of the previously tracked waypoint and the superscript GCS represents the frame in which the vector is expressed.
The components of the vector $\boldsymbol{r}^{GCS}$ when expressed in the frame  $\mathcal{X}_{ref}$ are  $x^p_e$ and  $y^p_e$ respectively.
This is achieved using a simple transformation between the frames GCS and $\mathcal{X}_{ref}$ seen in \eqref{eq:crosstrack_error}.

\begin{equation}
\boldsymbol{r}^{\mathcal{X}_{ref}} = R(\pi_p)^{\boldsymbol{T}} \boldsymbol{r}^{GCS}
\label{eq:crosstrack_error}
\end{equation}

The desired heading angle $\psi_d$ is the angle of the line joining $(x,y)$ and $(x_{LOS},y_{LOS})$ as measured from the global x-axis.
The desired heading angle is geometrically calculated as follows: 
\begin{equation}
    \psi_d = \pi_p - tan^{-1}\left(K_p^g y^p_e \right)      \label{LOS_traditional}
\end{equation}
where $K_p^g = 1/\Delta$. Adding an integral term that accumulates cross-track error with time compensates for steady-state error arising from environmental factors and unmodelled dynamics. Therefore, an integral component is incorporated to \eqref{LOS_traditional}, which results in the Integral LOS guidance scheme.
However, when a change in set-point occurs, the presence of an integral term leads to overshoots termed as integral windup. 
An alternative formulation that reduces this problem described in \cite{caharija2016integral} is employed as shown in \eqref{psi_desired} 
\begin{equation}
    \psi_d = \pi_p - tan^{-1}(K_p^g y^p_e +K_i^g y^p_{int}) \label{psi_desired}
\end{equation}
 where $y^p_{int}$ is a slowly varying substitute term whose dynamics are governed by the following equation
\begin{equation}
    \dot y^p_{int} = \frac{\Delta y^p_e}{\Delta^2 + (y^p_e + k y^p_{int})^2}
    \label{ype_int_dot}
\end{equation}
where $K_i^g = k\cdot K_p^g$ is the integral gain. In this study $k = 0.05$ is chosen. This scheme can be shown to render the equilibrium point $(y^p_e, y_{int}^p)=(0,0)$ globally asymptotically stable. For the Lyapunov stability proof, readers are referred to \cite{fossen2011handbook}.

\subsection{Way-point switching mechanism}
    
While tracking multiple waypoints, it is important to ascertain that the current waypoint is tracked when the vessel is in its vicinity. To consider whether a waypoint has been tracked or not, a circular region around the waypoint is considered. If the vessel enters this region, the waypoint is flagged as tracked. The criteria for switching waypoints is given by \eqref{eq:switch_waypoints}
\begin{equation}
    \sqrt{(y_{k+1} - y_{n})^2 + (x_{k+1} - x_{n})^2} \leq R_{tol}
    \label{eq:switch_waypoints}
\end{equation}

where $R_{tol}$ is the threshold value. In this study, this threshold is chosen to be thrice the ship length.

\subsection{Control Law}
\label{subsec:control}

The computed reference states from a guidance law are fed to the control system, which calculates the actuation commands necessary to converge the vessel along a desired trajectory with minimal deviation. This study uses a Proportional Derivative (PD) controller to ensure convergence of the vessel's heading $\psi$ to the desired heading angle $\psi_d$. The desired heading angle $\left(\psi_d\right)$ is computed using \eqref{psi_desired}. The error $e$ is defined as the difference between the current heading angle $\psi$ and this reference value, as shown in \eqref{Eq 18a}. The derivative of the error with respect to time is as shown in \eqref{Eq 18b}
\begin{subequations}        \label{Eq 18}
\begin{equation}
e = \psi - \psi_d     \label{Eq 18a}
\end{equation}
\begin{equation}
\dot e = \dot \psi - \dot \psi_d = r      \label{Eq 18b}
\end{equation}
\end{subequations}
 
where $\dot{\psi}_d$ is taken to be $0$ for the control implementation. This is valid as long as $\psi_d$ does not change faster than the time in which the controller can track the desired heading. The proportional derivative (PD) control law is expressed as
\begin{equation}
    \delta_c = -K_p^c e - K_d^c \dot e \label{PD_control}
\end{equation}
where $\delta_c$ is the rudder angle to be commanded, $e$ is the error in heading angle, and $\dot e$ is the rate of change of error. $K_p^c$ is the proportional gain, and $K_d^c$ is the derivative gain of the controller. These controller gains are to be tuned in such a way that deviation of the vessel's trajectory from the desired trajectory is minimal.
In this study, the controller gains $K_p^c$ and $K_d^c$ used are 3.5 and 4 respectively.

It is important to incorporate the practical limitations of the steering system. The absolute value of the rudder applied is saturated at $35^\circ$. The rudder rate when responding to a commanded rudder angle is assumed to follow a first-order dynamics subject to a saturation constraint, as shown in \eqref{eq:delta_dot_equation}

\begin{equation}
    \dot{\delta} = \begin{cases}
                   \frac{\delta_c - \delta}{T_\delta} & \text{, if } \left|\frac{\delta_c - \delta}{T_\delta}\right| \le \dot{\delta}_{max}\\
                    \text{sgn}\left(\frac{\delta_c - \delta}{T_\delta}\right) \dot{\delta}_{max} & \text{, otherwise }
               \end{cases}
    \label{eq:delta_dot_equation}
\end{equation}

where, $\delta_c$ is the commanded rudder angle, $\delta$ is the actual rudder angle, $T_\delta$ is the non-dimensional rudder rate time constant that is set to 1, and $\dot{\delta}_{max}$ is the nondimensional rudder rate saturation limit that is set to a value that corresponds to $5^\circ/\text{sec}$ for the full-scale ship. 

\subsection{Performance Metrics}
An optimal approach to track a desired goal location with obstacle avoidance entails minimizing the vessel's deviation from the original path while maintaining a lower control effort. Two metrics are defined to evaluate the performance of the guidance and control laws:
\begin{itemize}
    \item Controller Effort (CE) is defined as the normalized mean value of the rudder applied over the trajectory and is mathematically defined as
        \begin{equation}
            CE = \frac{1}{\delta_{max} \cdot T}\int_{0}^{T} \left|\delta(\tau)\right| d\tau
            \label{eq:controller_effort_equation}
        \end{equation}
        where $T$ corresponds to the total time of the trajectory.
        The maximum rudder angle $\delta_{max}$ that can be applied in this study is chosen as $35^\circ$. Note that $\delta$ is the actual rudder angle being applied at a particular time and is different from the commanded rudder angle $\delta_c$.  
            
    \item Mean Cross Track Error (MCTE) mathematically quantifies the closeness of the actual trajectory to the desired trajectory and is defined as shown in \eqref{eq:crosstrack_metric_equation}. 
        \begin{equation}
            MCTE = \frac{1}{L\cdot T}\int_{0}^{T} \left|{y}_{e}^{p}(\tau)\right| d\tau
            \label{eq:crosstrack_metric_equation}
        \end{equation}
        Note that the metric has been non-dimensionalized with respect to the ship length $L$.
        
\end{itemize}
The CE metric provides a quantifiable measure of the control input applied to the vessel during the tracking process. The MCTE metric serves as an indicator of the vessel's deviation from the original trajectory. By considering both the CE and MCTE metrics, we can comprehensively evaluate and optimize the vessel's tracking capabilities, enabling efficient goal attainment while effectively navigating obstacles.

\subsection{Results}
\begin{figure*}[htbp]
    \centering
    \begin{subfigure}{\textwidth}
      \centering
      \includegraphics[scale = 0.75]{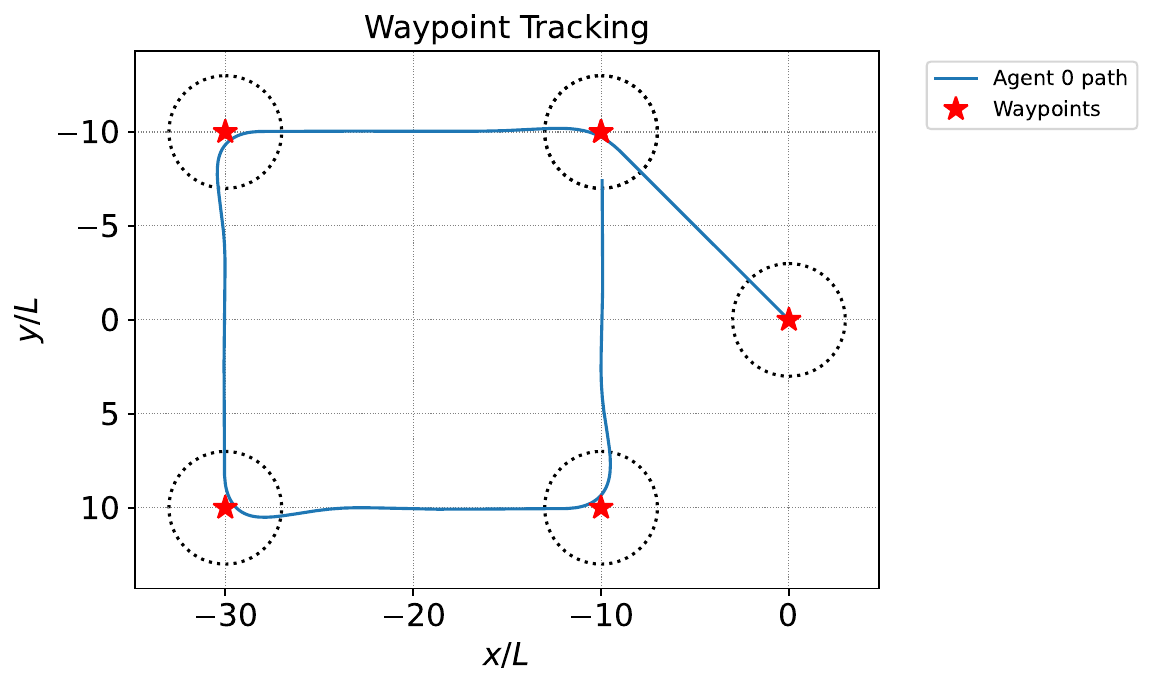}
        \caption{Vessel performing waypoint tracking based on the parameters in \autoref{tab:gnc_params_in_wptracking}. 
            Four waypoints are provided. The dotted circle represents the radius of tolerance that is set to three ship lengths in this study.}
     \label{fig:waypoint_tracking_individual_result}
    \end{subfigure}
    
    \begin{subfigure}{0.45\textwidth}
      \centering
      \includegraphics[width=1\linewidth]{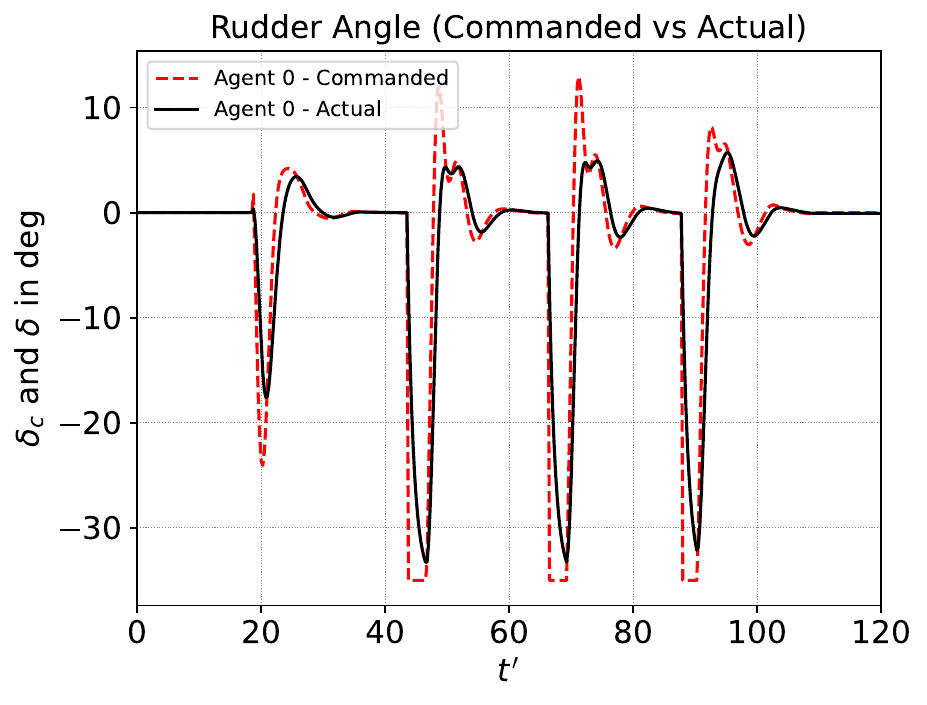}
      \caption{Commanded vs. Actual Rudder Angle. Controller effort defined in \eqref{eq:controller_effort_equation} comes out as $\text{CE} = 0.1203$. }
      \label{fig:waypointtracking_commandedvsactual_rudder}
    \end{subfigure}
    \hfill
    \begin{subfigure}{0.45\textwidth}
      \centering
      \includegraphics[width=1\linewidth]{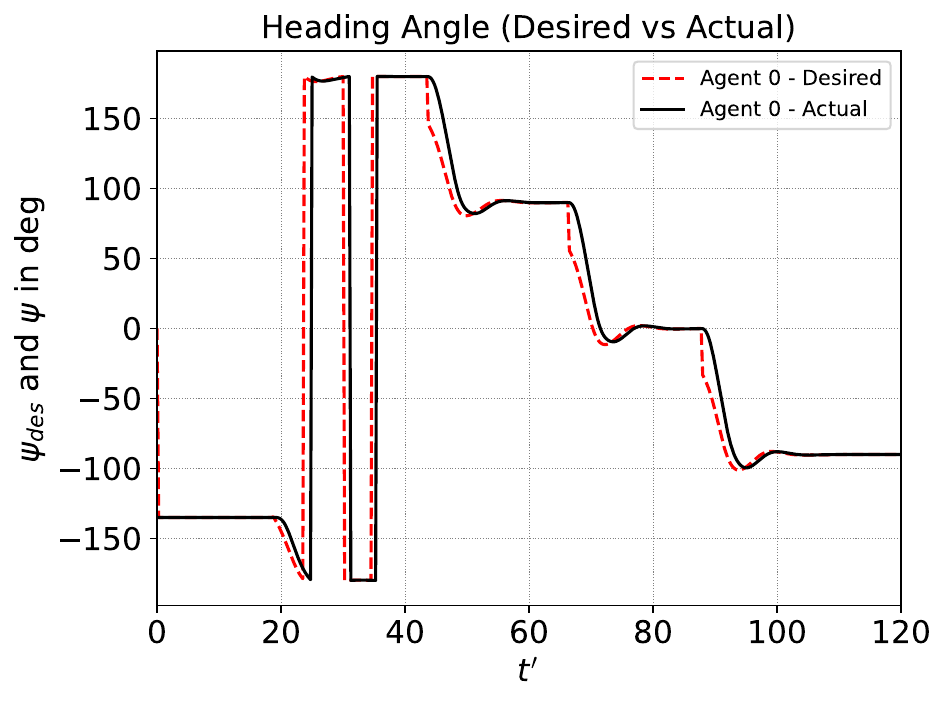}
      \caption{Desired Heading angle vs. Actual Heading angle of the ASV while tracking the waypoints.}
      \label{fig:waypoint_tracking_heading_angle}
    \end{subfigure}
    \hfill
    \begin{subfigure}{0.45\textwidth}
      \centering
      \includegraphics[width=1\linewidth]{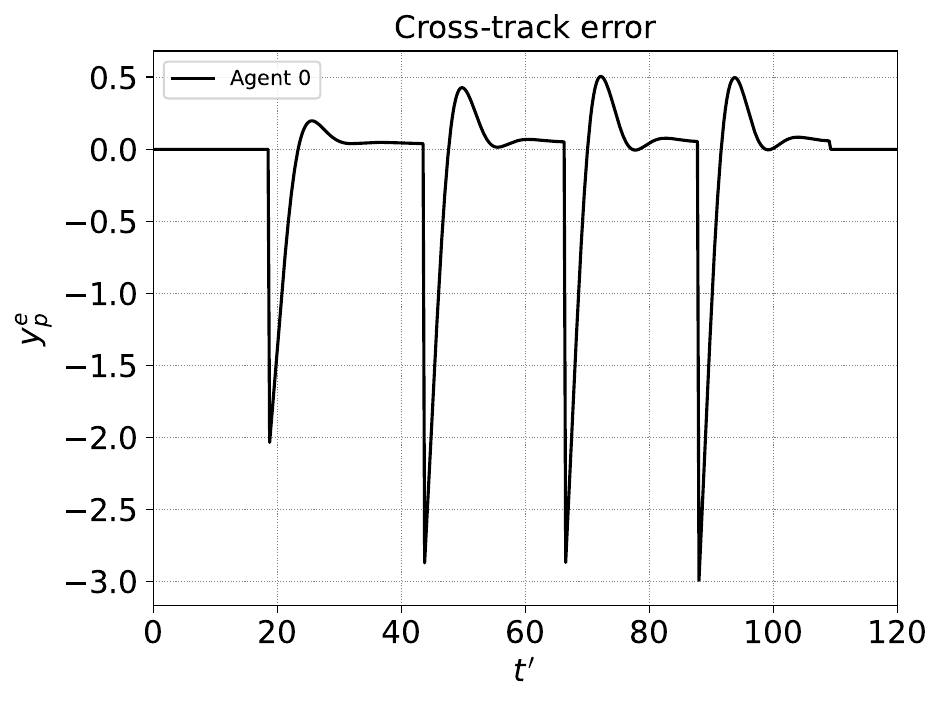}
      \caption{Variation of the Cross-track Error. Mean cross-track error defined in \eqref{eq:crosstrack_metric_equation} comes out as $\text{MCTE} = 0.2655L$.}
      \label{fig:waypoint_tracking_crosstrackerror}
    \end{subfigure}
    
    \caption{Simulation Results of Waypoint Tracking}
    \label{fig:waypoint_tracking_results}
\end{figure*}

Based on the above description of guidance and control mechanism, a simulation-based study is conducted based on the numerical model described in \eqref{eq:mmg_nd}.
A set of waypoints forming a square shape is given, and the vessel is tasked to track the waypoints.
\autoref{tab:gnc_params_in_wptracking} describes the set of guidance and control parameters used for performing waypoint tracking. 
\autoref{fig:waypoint_tracking_results} shows the path followed during waypoint tracking, rudder angle versus time, heading angle versus time, and the variation of cross-track error with time. 
The overshoots after tracking the waypoints as observed in \autoref{fig:waypoint_tracking_crosstrackerror} can be attributed to the sluggish dynamics of the vessel that are evident from \autoref{fig:waypoint_tracking_heading_angle} where a slight delay between the desired heading and actual heading can be observed.


\begin{table}[htbp]
    \renewcommand{\arraystretch}{1.2}
    \caption{Guidance and Control Parameters\\
    Note: All the quantities with \textit{L} are non-dimensional in ship-lengths}
    \begin{tabular*}{\tblwidth}{@{} LL@{} }
     \toprule
      \textbf{Parameter} & \textbf{Value}\\
     \midrule
      $K_p^c$ & 3.5\\
      $K_d^c$ & 4\\
      $K_p^g = 1/ \Delta$ & 0.5\\
      $K_i^g$ & 0.025\\
      $\Delta$ & 2 (L)\\
      $R_{tol}$ & 3 (L)\\
     \bottomrule
    \end{tabular*}
    \label{tab:gnc_params_in_wptracking}
  \end{table}


\section{Reactive Guidance - Artificial Potential Fields (APFs)}
\label{sec:apfs}
So far, the formulation described above allows the vessel to track different waypoints using the ILOS guidance and PD control law. However, when static or dynamic obstacles are encountered in the environment the vessel must take recourse to a reactive guidance that takes proactive steps to avoid collision. This study investigates the effectiveness of artificial potential fields (APFs) for this reactive guidance. 

The APF algorithm is based on the concept of potential fields, which is a mathematical construct that models the behavior of objects in a physical system as if they were subject to virtual forces that pull them toward the goal location and push them away from the obstacles. The potential field is calculated by the scalar addition of its two components, the attractive and repulsive potential. The potential is defined at every point in the configuration space. It has a higher value when the vessel is near an obstacle and is away from the goal. It has a lower value when the vessel is farther away from obstacles and closer to the goal. The potential function also has its lowest value (global minima) at the goal location, and the value increases for the configurations that are further away from the goal. The agent moves along the direction of the steepest descent of the generated potential field that is calculated by its gradient.

\subsection{Inverse Square Potential}

\begin{figure}[htbp]
    \centering
    \includegraphics[width=\linewidth]{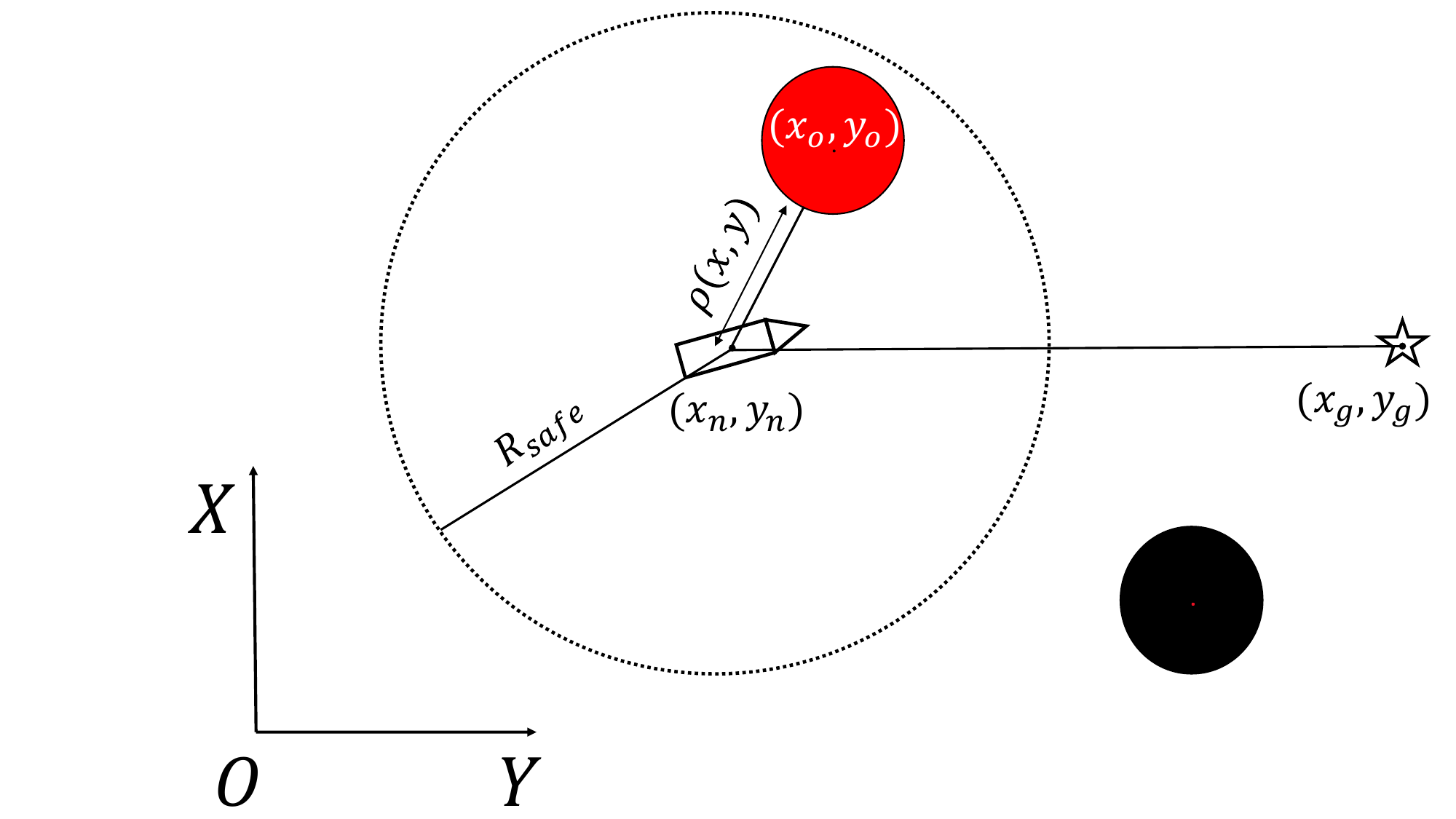}
    \caption{Encounter Scenario}
    \label{fig:encounter_scene}
\end{figure}

\begin{figure*}[htbp]
    \centering
    \includegraphics[width= 0.9\linewidth]{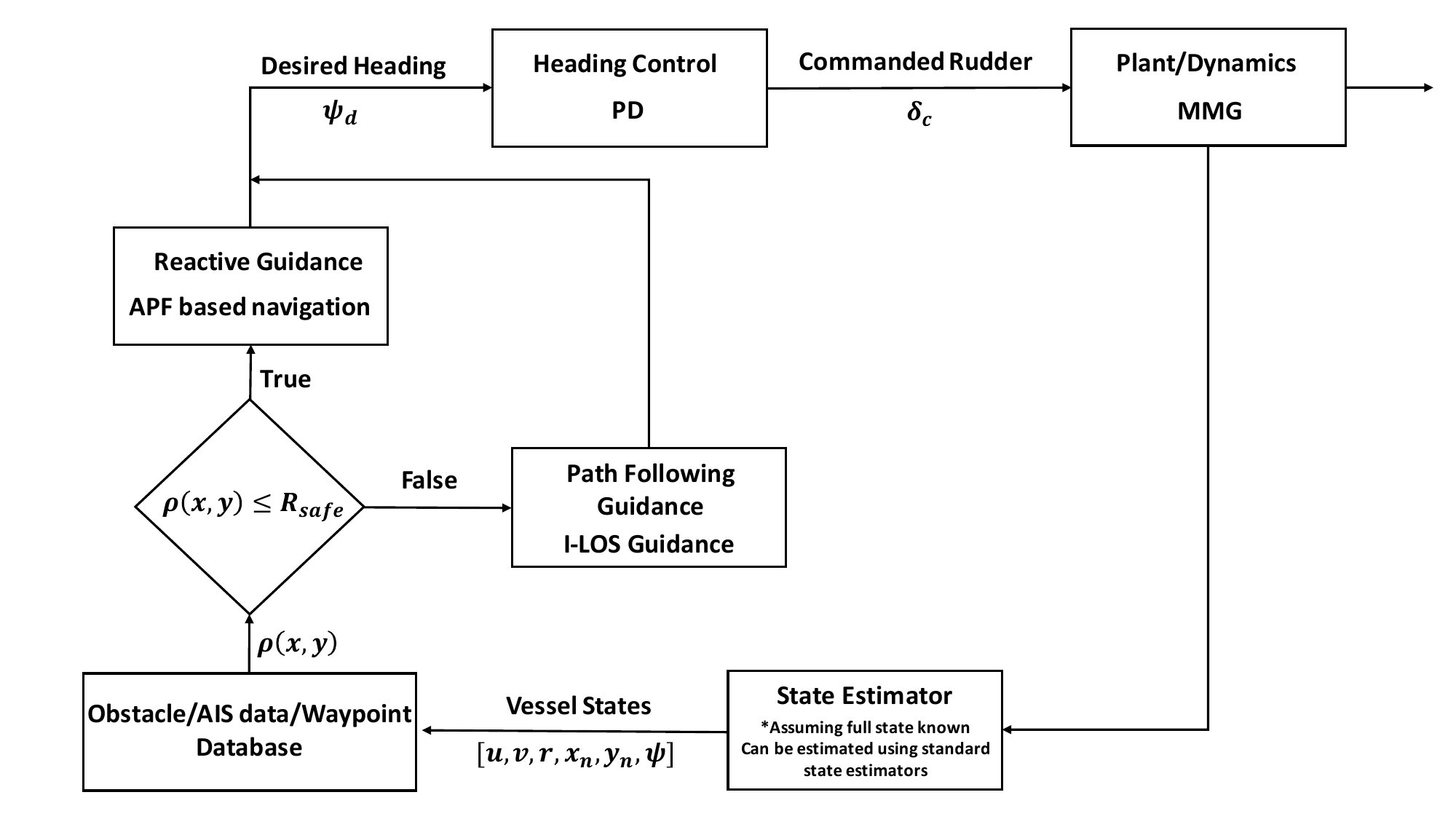}
    \caption{Guidance, Navigation, and Control Block Diagram}
    \label{fig:gnc_block_diagram}
\end{figure*}

The inverse square potential is one of the oldest potential functions proposed for reactive guidance. It was originally introduced by \cite{khatib1986real} for robotic manipulators. The attractive potential function, $\phi_{a}(x)$ is constructed by using the distance between the current position of the robot, $\boldsymbol{X_n} = \begin{bmatrix} x_{n}& y_{n} \end{bmatrix}^T$ and the goal location $\boldsymbol{X_g} = \begin{bmatrix} x_{g} & y_{g} \end{bmatrix}^T$ as shown in \eqref{eq:attractive_potential}
\begin{equation}
    \phi_{a}(x) = k_{att} \left|\left|x - x_{g}\right|\right|_{2}^2
    \label{eq:attractive_potential}
\end{equation}
where $k_{att}$ is a constant scaling parameter and $\left|\left|.\right|\right|_{2}$ represents two-dimensional Euclidean norm. The repulsive potential function $\phi_{r}(x)$ is defined using the distance function, $\rho(x)$, and is expressed as shown in \eqref{eq:repulsive_potential}
\begin{equation}
    \phi_{r}(x) = \begin{cases}
                    k_{rep} \left(\frac{1}{\rho(x)} - \frac{1}{d_{0}}\right)^{2} & \text{, if } \rho(x) \le d_{0}\\
                    0 & \text{, if } \rho(x) > d_{0} 
               \end{cases}
    \label{eq:repulsive_potential}
\end{equation}
where $k_{rep}$ is a constant scaling parameter and $d_{0}$ is a scalar that controls the influence of the obstacle field. The distance $\rho(x,y)$ can be computed as seen in \eqref{eq:rho_xy}, where $R_{obs}$ represents the radius of the obstacle.
\begin{equation}
    \rho(x,y) = \sqrt{\left(x_n - x_o\right)^2 + \left(y_n - y_o\right)^2} - R_{obs}
    \label{eq:rho_xy}
\end{equation}
The total potential function can be obtained using a scalar addition of the attractive and repulsive fields as seen in \eqref{eq:combined_potential}.
\begin{equation}
    \phi(x) = \phi_{a}(x) + \phi_{r}(x)
    \label{eq:combined_potential}
\end{equation}
Since the analytical expression of the function is known and the potential is defined over the continuous space, the gradient of the function can be analytically evaluated. Consider an encounter scenario as seen in \autoref{fig:encounter_scene} where a static obstacle is present. A single static obstacle in the red is present at $\boldsymbol{X_o} = \begin{bmatrix} x_{o} & y_{o} \end{bmatrix}^{\boldsymbol{T}}$, while $\boldsymbol{X_n} = \begin{bmatrix} x_{n} & y_{n} \end{bmatrix}^{\boldsymbol{T}}$ and $\boldsymbol{X_g} = \begin{bmatrix} x_{g} & y_{g} \end{bmatrix}^{\boldsymbol{T}}$ represent the current location of the ship and the goal location respectively.
The parameter $R_{safe}$ denotes the distance from the center of the vessel at which the vessel will consider the obstacle detected. For obstacles within this radius $R_{safe}$, the ship will have a repulsive potential associated with them, while the others will have zero repulsive potential. 
Once an obstacle is found to be within $R_{safe}$ distance of the vessel, the reactive APF based guidance will supersede the path following ILOS guidance module. \autoref{fig:gnc_block_diagram} explains the overall navigation scheme in the presence of obstacles in the environment.

\begin{figure}[htbp]
    \centering
    \includegraphics[width=\linewidth]{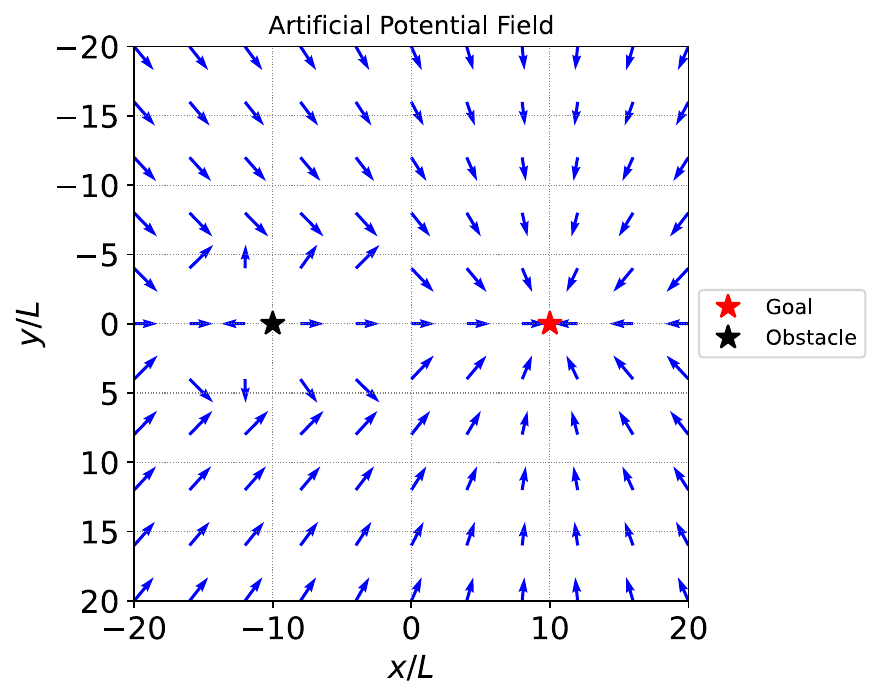}
    \caption{Inverse Square Vector-field}
    \label{fig:streamlines_inv_square}
\end{figure}

\begin{figure*}[htbp]
    \centering
    \begin{subfigure}{\textwidth}
      \centering
      \includegraphics[width=\textwidth]{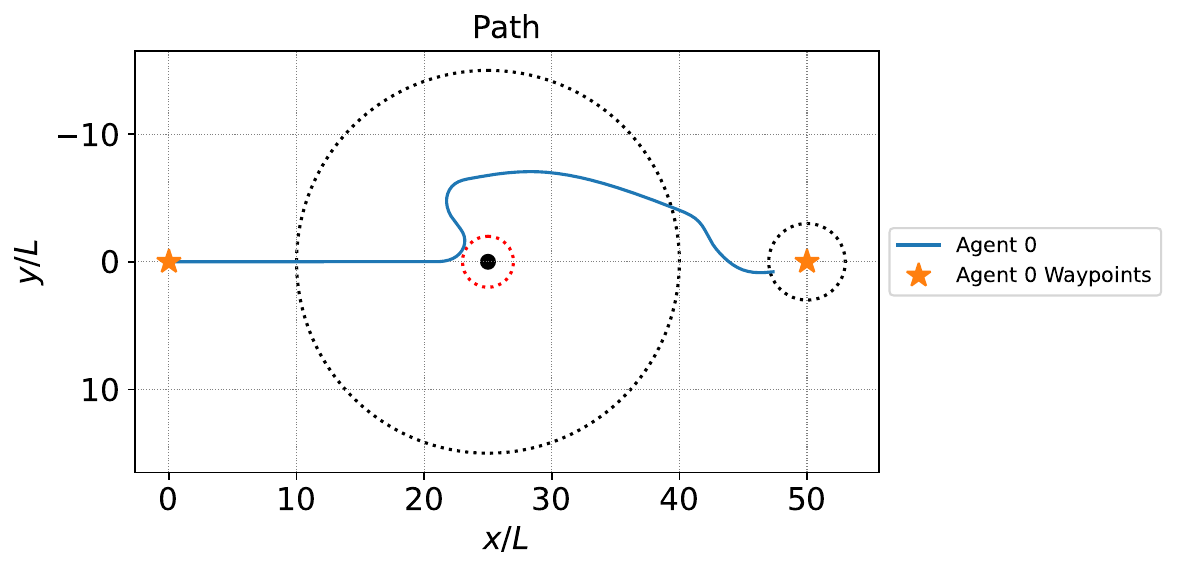}
      \caption{Vessel Trajectory: Large Overshoots observed while avoiding obstacles can be attributed to sluggish vessel dynamics and the way the potential is formulated.
                The vessel approaching head-on to the obstacle is not commanded to change the heading until it reaches the stagnation point as is evident from \autoref{fig:streamlines_inv_square}.}
      \label{fig:inverse_square_trajectory}
    \end{subfigure}
    \begin{subfigure}{0.45\textwidth}
      \centering
      \includegraphics[width=\textwidth]{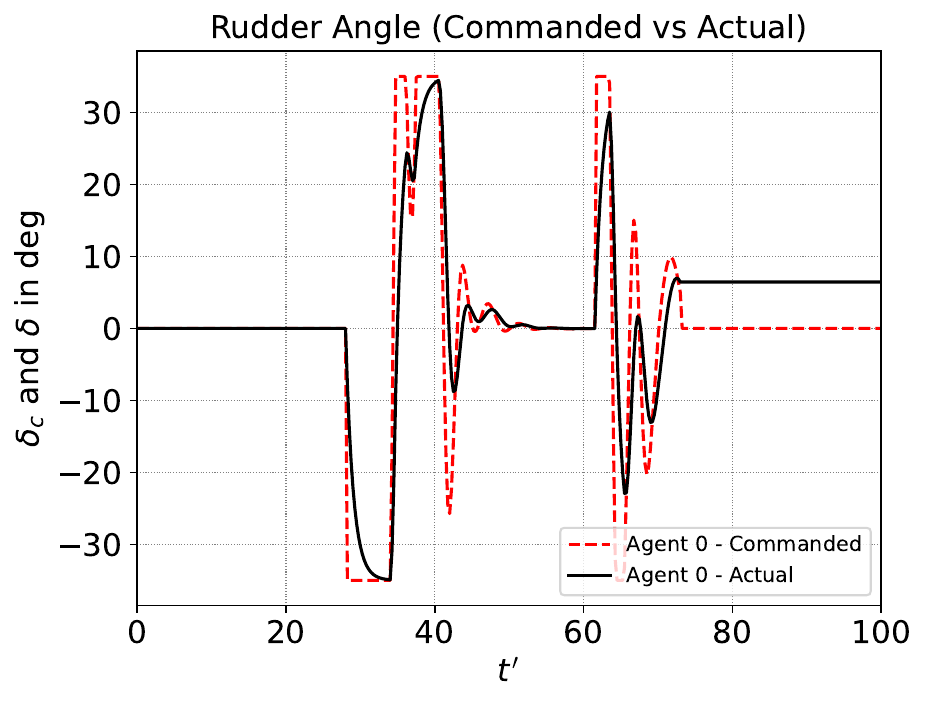}
      \caption{Comparison of Commanded Rudder Angle vs. True Rudder Angle. Evaluated Controller Effort $\text{(CE)} = 0.2015$}
      \label{fig:inverse_square_rudder}
    \end{subfigure}
    \hfill
    \begin{subfigure}{0.45\textwidth}
      \centering
      \includegraphics[width=\textwidth]{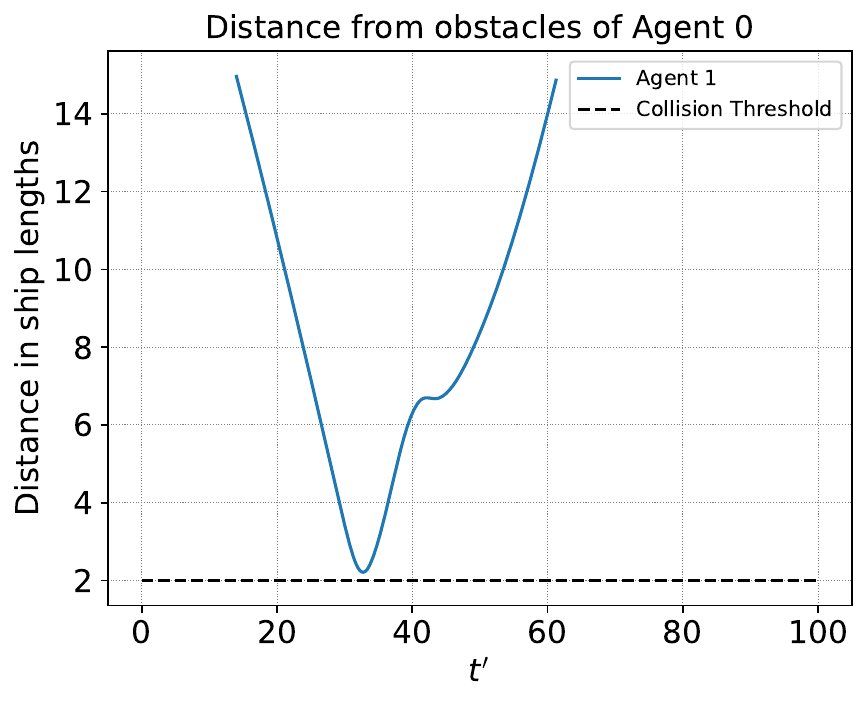}
      \caption{Variation of distance to obstacle over the maneuver. Agent 1 refers to the static obstacle in this case}
      \label{fig:inverse_square_obs_dist}
    \end{subfigure}
    
    \caption{Obstacle Avoidance using Inverse Square Potential Function }
    \label{fig:inverse_square_main}
\end{figure*}

Using \eqref{eq:attractive_potential} and \eqref{eq:repulsive_potential}, the gradient of the total potential function can be computed as shown in \eqref{eq:grad_phi_inv_square}.
\begin{equation}
    \begin{aligned}
    -\boldsymbol{\nabla} \phi &=  -k_{att} \left(\boldsymbol{X_n} -\boldsymbol{X_g}\right) \\
    & + \frac{2 k_{rep} \left(\boldsymbol{X_n}-\boldsymbol{X_o}\right)}{\left(\rho\left(x,y\right)\right)^2 \left(\rho\left(x,y\right) + R_{obs}\right)}  \left(\frac{1}{\rho\left(x,y\right)} - \frac{1}{d_o}\right) 
    \end{aligned}
    \label{eq:grad_phi_inv_square}
\end{equation}
%
The direction of the vector $-\boldsymbol{\nabla} \phi$ from \eqref{eq:grad_phi_inv_square} represents the desired heading direction that the vessel must maintain to avoid obstacles and reach the goal location.
\autoref{fig:streamlines_inv_square} shows the plot of the gradient field described by \eqref{eq:grad_phi_inv_square}, where it can be seen that the field repels the vessel away from the obstacle at $\boldsymbol{X_o} = \begin{bmatrix} -10 & 0 \end{bmatrix}^{\boldsymbol{T}}$ and attracts towards the goal location at $\boldsymbol{X_g} = \begin{bmatrix} 10 & 0 \end{bmatrix}^{\boldsymbol{T}}$. It can be observed from \autoref{fig:streamlines_inv_square} that while the effect of the goal is felt globally, the effect of the obstacle is perceived only locally. The parameters $k_{rep} \text{ and } k_{att}$ chosen here are 200000 and 50 respectively.
%
Using the dynamics described in \eqref{eq:mmg_nd} and the theory described above, the KCS vessel is simulated in an environment with a single static obstacle located in the path of the vessel using the inverse square APF as the repulsive potential. The results are shown in \autoref{fig:inverse_square_main}.
The vessel starts from the origin with an initial heading of $0^\circ$ with a waypoint to be tracked at $\left[50, 0\right]$, and the obstacle is located at $\left[25, 0\right]$. The APF parameters are reported in \autoref{tab:inverseapfparameters}. Note that the simulation is performed for a safe radius of $R_{safe} = 15L$. In order to show the portion of the vessel's path where the presence of the obstacle played a role through the reactive guidance, a dotted circle around the obstacle is shown in \autoref{fig:inverse_square_trajectory}.
\begin{table}[htbp]
    \renewcommand{\arraystretch}{1.2}
    \caption{APF parameters}
    \begin{tabular*}{\tblwidth}{@{} LL@{} }
    \toprule
    \textbf{Parameter} & \textbf{Value} \\
    \midrule
        $k_{rep}$ & 200000\\
        $k_{att}$ & 50\\
    \bottomrule
   \end{tabular*}
    \label{tab:inverseapfparameters}
\end{table}  

As seen in \autoref{fig:inverse_square_trajectory}, the vessel trajectory deviates up to $7L$ from the expected path without the obstacle. It can be seen from \autoref{fig:inverse_square_rudder} that the rudder is not applied until $t'=25$ and by this time it is seen from \autoref{fig:inverse_square_obs_dist} that the vessel is less than $6L$ away from the obstacle. This behavior can be attributed to the way the potential field is formulated, especially when the vessel is on a collision course with the obstacle.
A stagnation point can be seen in \autoref{fig:streamlines_inv_square} where the direction of the gradient changes at around $\begin{bmatrix} -15 & 0 \end{bmatrix}$, which is merely $5L$ from the obstacle. Thus, even though APF is activated when distance between vessel and obstacle is $15L$, the gradient reverses only about $5L$ from the obstacle. The vessel is headed toward the obstacle until it reaches this stagnation point. Since the vessel dynamics are non-holonomic in nature, tracking sudden changes in the desired heading angle due to sharp changes in gradient becomes difficult. Thus the vessel comes close to collision threshold of $2L$ before making a large deviation from the original course.

\begin{figure}
    \centering
    \includegraphics[width=0.5\textwidth]{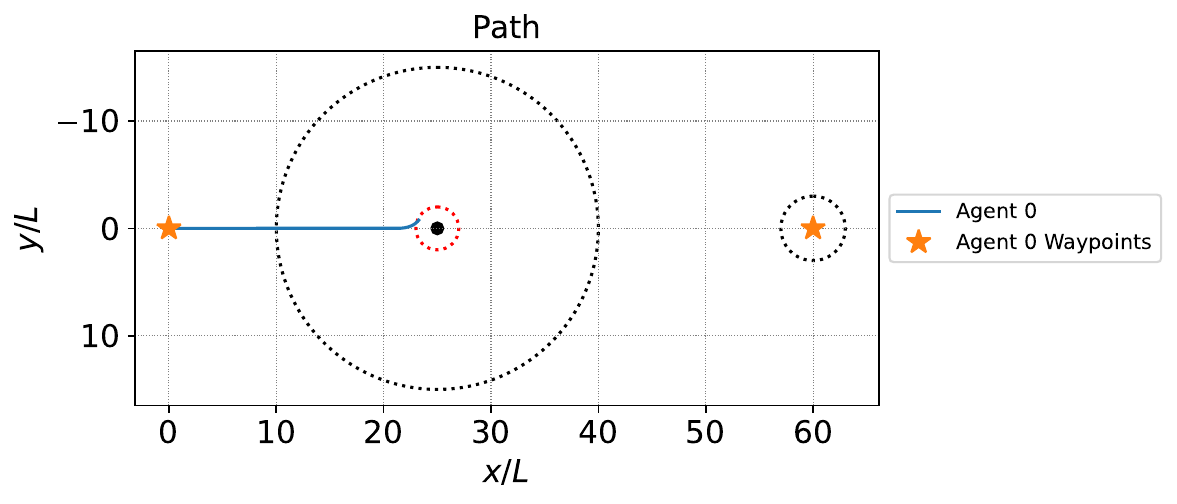}
    \caption{Collision due to gradients due to goal being larger than gradients due to obstacles}
    \label{fig:inverse_square_failure}
\end{figure}

Another problem with the inverse square APF is linear dependence of the gradient on the distance of vessel from the obstacle and the goal. Thus, the goal being farther away may cause the effect of closer obstacles to be dwarfed in the gradient computation. This is demonstrated in \autoref{fig:inverse_square_failure} where the vessel is tasked to track the waypoint $\left[60L, 0\right]$ instead of $\left[50L, 0\right]$ as shown in \autoref{fig:inverse_square_main}. Note that obstacle location is fixed at $\left[25L, 0\right]$ in both cases. However, it is seen from \autoref{fig:inverse_square_failure} that the vessel enters the collision threshold in this case and is unable to reach the goal waypoint. 

In addition to this, the inverse square APF is also known to have other limitations. One of these is that the local minima in this formulation does not occur exactly at the goal but is some distance away from the goal \citep{kim1992real}. This paves the way for alternative potential functions to be explored. In order to overcome the problem of local minima and sudden changes in the gradients (stagnation points), harmonic functions \citep[originally proposed by][]{kim1992real} are explored for collision avoidance in the next section. 

\subsection{Harmonic Potential Functions}
\label{sec:Harmonic Functions}
In fluid dynamics, harmonic functions are used to describe the velocity potential of a fluid, which is a scalar field that satisfies the Laplace equation shown in \eqref{eq:Laplace equation}
\begin{equation}
    \nabla^{2}\phi  = 0
    \label{eq:Laplace equation}
\end{equation}
where $\nabla^2 \equiv \boldsymbol{\nabla}\cdot\boldsymbol{\nabla}$ is called the Laplacian Operator and $\phi$ is the velocity potential. Harmonic functions are solutions of the Laplace equation and are smooth, continuous, and differentiable throughout the domain. 

\begin{figure}[htbp]
    \centering    
    \includegraphics[width=0.5\linewidth]{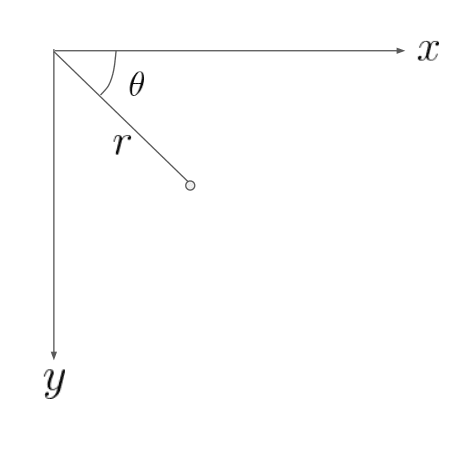}
    \caption{Coordinate frame with z-axis pointed into the plane - positive orientation of $\theta$ is clockwise from x-axis}
    \label{fig:polar_coordinate frame}
\end{figure}

In this study, specifically, a set of combinations of some original harmonic functions like source, sink and vortex are explored. These functions depend on $r$ (distance from the origin) and $\theta$ (orientation from the x-axis) as seen in \autoref{fig:polar_coordinate frame}.
To compute the flow velocity $\boldsymbol{V} = \begin{bmatrix} v_r & v_\theta \end{bmatrix}^T$, $v_r$ and $v_\theta $ can be computed as seen in \eqref{eq:vr vtheta equation}.
\begin{equation}
    v_r = \frac{\partial \phi}{\partial r} \qquad v_\theta = \frac{1}{r}\frac{\partial \phi}{\partial \theta}
    \label{eq:vr vtheta equation}
\end{equation}
Two fundamental flows are summarized below.
\begin{itemize}
    \item Source/Sink: \\
    In two dimensions, a source/sink at the origin can be represented by:
    \begin{equation}
        \phi (r, \theta)  = \frac{\Lambda}{2\pi} \log(r)
        \label{eq:Source/Sink equation}
    \end{equation}
    The magnitude of $\Lambda$ is called the strength of the source ($\Lambda > 0$) or the sink ($\Lambda < 0$). 
    The flow velocity in polar coordinates can be obtained using \eqref{eq:vr vtheta equation} as:
    \begin{equation}
        \boldsymbol{V} = \begin{bmatrix} \frac{\Lambda }{2 \pi  r} & 0\end{bmatrix}^T
        \label{eq:Vsink equation}
    \end{equation}    
    \item Vortex \\
    A vortex is a circulating steady motion, where $v_r = 0$ and $v_\theta = f(r)$ (depends only on the radial distance from the center).
    In two dimensions, a vortex at the origin can be represented by:
    \begin{equation}
        \phi (r, \theta)  = \frac{K}{2\pi} \theta
        \label{eq:Vortex equation}
    \end{equation}
    A positive sign of $K$ results in a clockwise vortex while a negative sign results in a counter-clockwise vortex.
    The flow velocity can be obtained using \eqref{eq:vr vtheta equation} as:
    \begin{equation}
        \boldsymbol{V} = \begin{bmatrix} 0&\frac{K}{2  \pi r}\end{bmatrix}^T
        \label{eq:Vvortex equation}
    \end{equation}
\end{itemize}

\begin{figure}
    \centering
    \includegraphics[width=\linewidth]{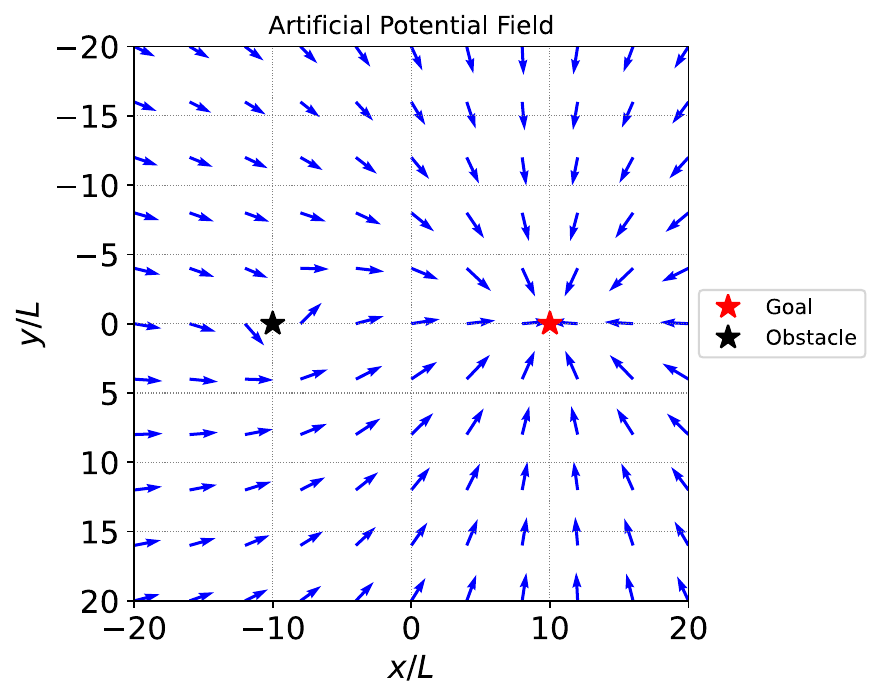}
    \caption{Vector-field due to sink at goal and vortex center at obstacle}
    \label{fig:streamlines_vortex}
\end{figure}

This study proposes the combinations of the above-described harmonic functions to model a flow around an obstacle. The sink is used to represent the goal location which causes the attractive potential. The obstacle is represented by a vortex, which causes the repulsive potential. The vector field developed by this combination is shown in \autoref{fig:streamlines_vortex}.

Unlike the inverse square APF, where the gradients are strongly affected by the vessel distance to obstacle or goal, the harmonic potential field gradients decay in strength away from obstacle or goal. Thus, this approach allows closer obstacles to have more weight on the collision avoidance maneuver as compared to the far away obstacles and goal. 

\begin{figure*}[htbp]
    \centering
    \begin{subfigure}{\textwidth}
      \centering
      \includegraphics[width=\textwidth]{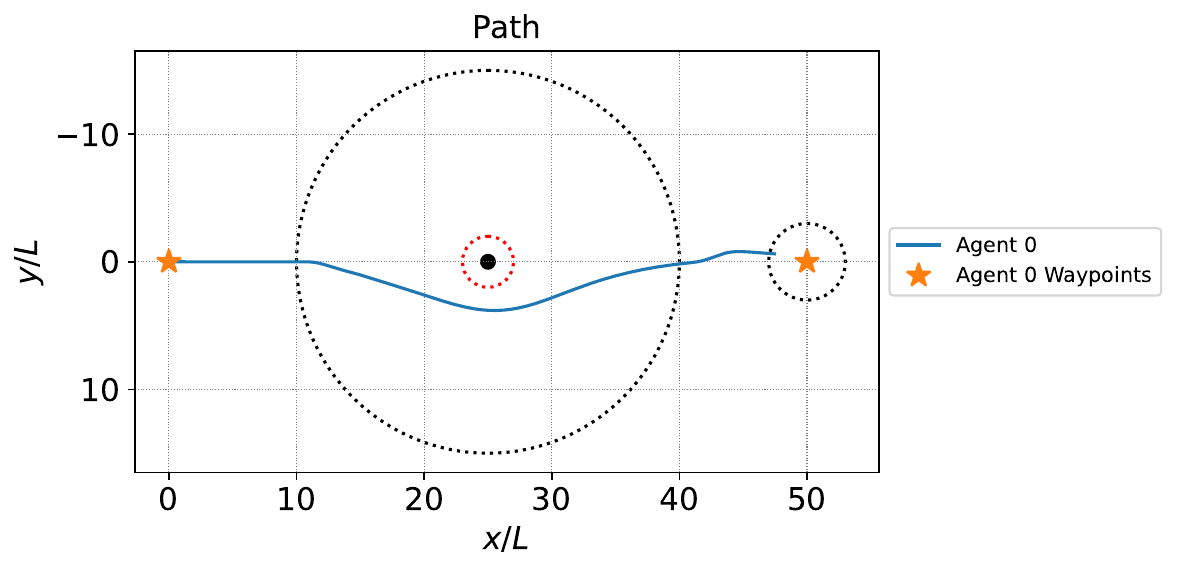}
      \caption{Vessel Trajectory: Smaller overshoots are observed while avoiding obstacles as the heading changes smoothly once the obstacle is within the radius of $15L$ of the vessel}
      \label{fig:vortex_static_obs_path}
    \end{subfigure}
    \begin{subfigure}{0.45\textwidth}
      \centering
      \includegraphics[width=\textwidth]{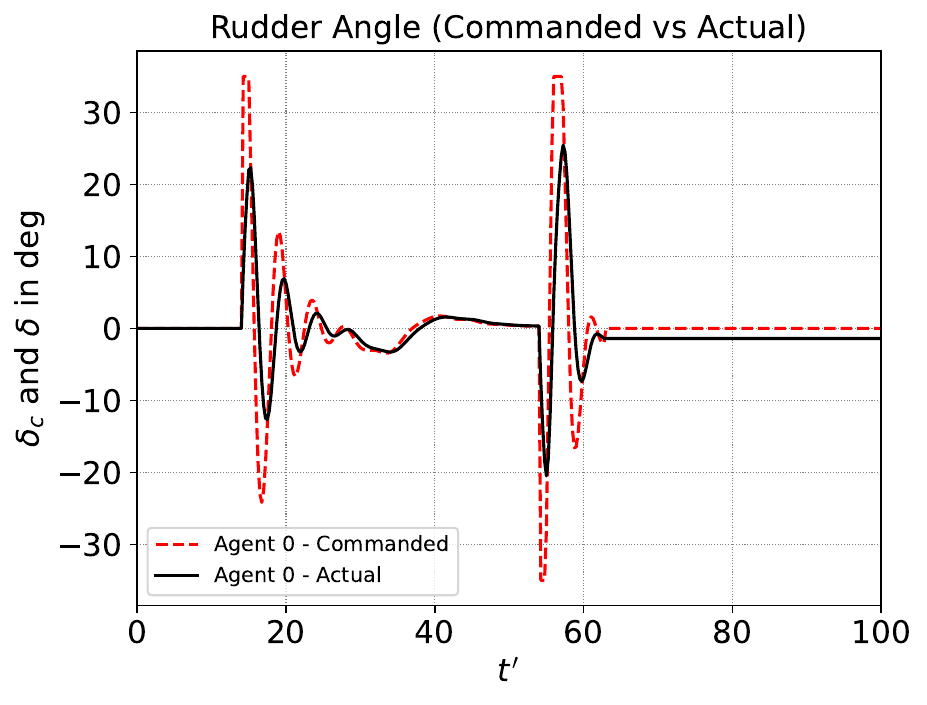}
      \caption{Comparison of Commanded Rudder Angle vs. True Rudder Angle. Evaluated Controller Effort $\text{(CE)} = 0.0852$}
      \label{fig:vortex_static_obs_rudder}
    \end{subfigure}
    \hfill
    \begin{subfigure}{0.45\textwidth}
      \centering
      \includegraphics[width=\textwidth]{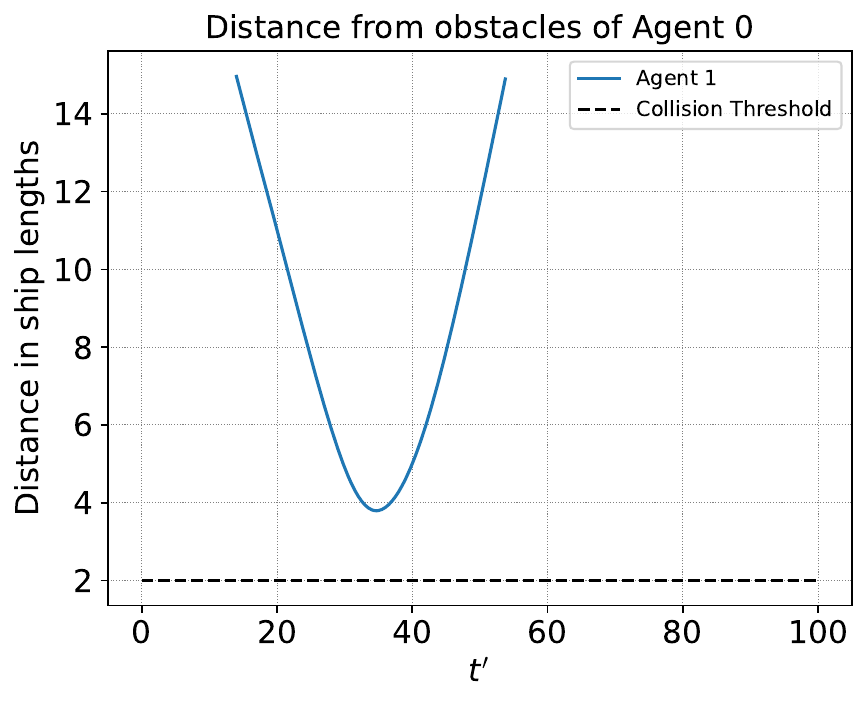}
      \caption{Variation of distance to obstacle over the maneuver. Agent 1 refers to the static obstacle in this case}
      \label{fig:vortex_static_obs_dist}
    \end{subfigure}
    
    \caption{Obstacle Avoidance using Sink-Vortex Potential Function with $K_{vor}=-10$ and $\Lambda_{sink}=-100$}
    \label{fig:vortex_static_obs_main}
\end{figure*}

An example of the sink-vortex approach to track a goal waypoint in the presence of static obstacle is shown in \autoref{fig:vortex_static_obs_main}. Note that this is the same case as investigated with inverse square potential in \autoref{fig:inverse_square_main}. Unlike the inverse square case, the rudder is applied almost immediately after sensing the obstacle in $R_{safe}=15L$. It is also seen that the minimum distance between the vessel and obstacle is almost twice as much observed in the inverse square APF case. The control effort as seen for the rudder angle plots is also much lower than for the inverse square APF case.

\begin{figure}
    \centering
    \includegraphics[width=\linewidth]{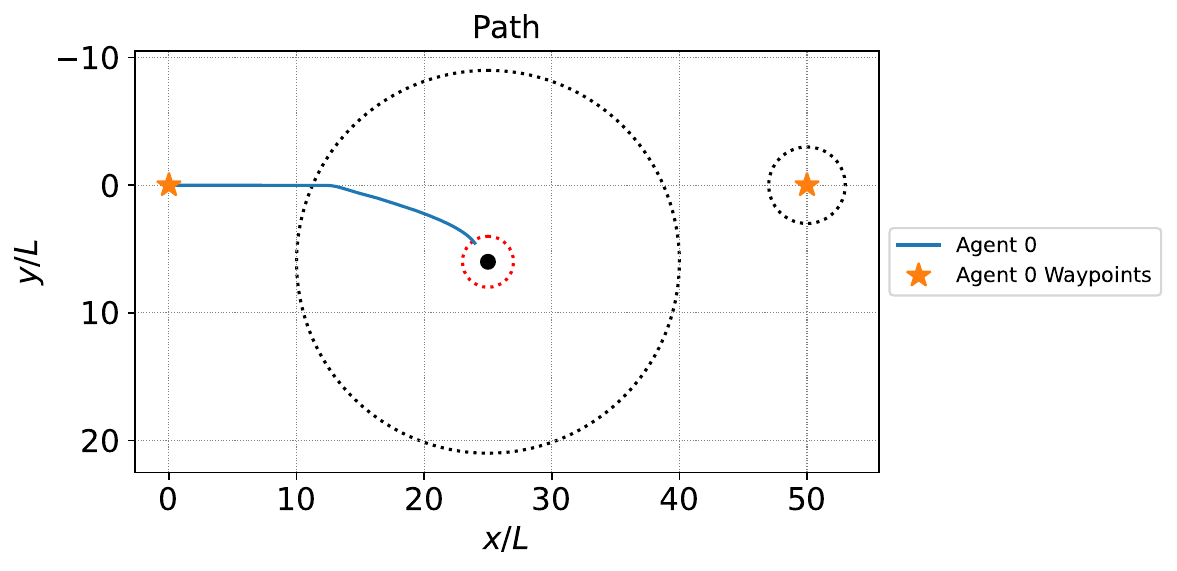}
    \caption{Case where the sink-vortex approach fails}
    \label{fig:vortex_failure}
\end{figure}

\begin{figure}[htbp]
    \centering
    \includegraphics[width=\linewidth]{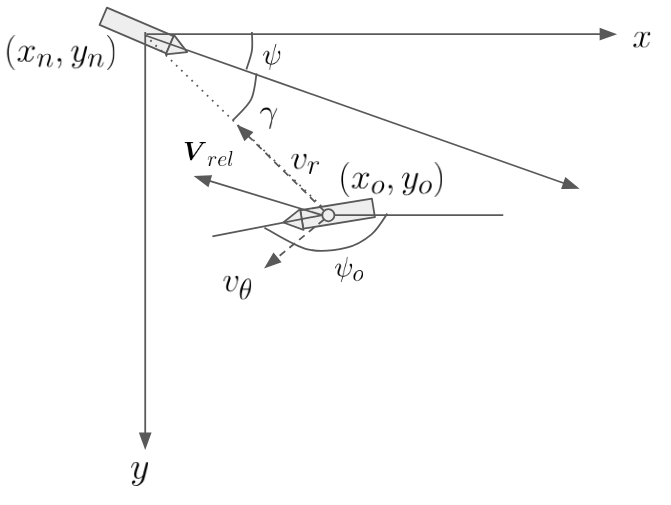}
    \caption{Modification to vortex at obstacle}
    \label{fig:gamma}
\end{figure}

Although this sink-vortex approach overcomes the limitations of the inverse square potential, it too has a limitation. When the vessel approaches a obstacle on the starboard side, it is drawn towards the obstacle and may at times lead to collision. This is illustrated in \autoref{fig:vortex_failure}. Thus the sink-vortex approach needs a modification to overcome this limitation.
%
%
The vortex approach is modified to be active only when the relative tangential velocity of the obstacle with respect to the vessel is positive and the obstacle lies in front of it. Consider the scenario shown in \autoref{fig:gamma}. The angle $\gamma$ defined as the angle between the line joining the vessel and obstacle and the vessel's bow can be calculated as shown in \eqref{eq:gamma}
\begin{align}
    \gamma = \tan^{-1} \left(\frac{y_o - y_n}{x_o - x_n}\right) - \psi
    \label{eq:gamma}
\end{align}
where the vessel position is given by $(x_n, y_n)$ and the obstacle position is given by $(x_o, y_o)$. The relative velocity of the obstacle expressed in GCS is given by 
\begin{align}
    \boldsymbol{V}_{rel} = -R(\psi)\boldsymbol{\nu} = R(\psi)\begin{bmatrix} -u \\ -v \end{bmatrix}
    \label{eq:rel_vel}
\end{align}
In case of dynamic obstacle, this is replaced by
\begin{align}
    \boldsymbol{V}_{rel} = R(\psi_o)\boldsymbol{\nu}_{o} -R(\psi)\boldsymbol{\nu}
    \label{eq:rel_vel}
\end{align} 
where $\boldsymbol{\nu}_{o} = \begin{bmatrix} u_{o} & v_{o} \end{bmatrix}^{\boldsymbol{T}}$ is the velocity of the dynamic obstacle expressed in BCS of obstacle and $\psi_o$ is the heading angle of the obstacle with respect to x-axis of GCS. The radial and tangential relative velocity can be obtained by
\begin{align}
    \begin{bmatrix}
        v_{r} \\
        v_{\theta}
    \end{bmatrix} = \begin{bmatrix}
        \cos(\gamma) & \sin(\gamma) \\
        -\sin(\gamma) & \cos(\gamma) \\
    \end{bmatrix} \boldsymbol{V}_{rel}
\end{align}

The modified vortex strength $K_{vor}$ is given by
\begin{align}
    K_{vor} = \begin{cases}
        0 \quad \text{ if } v_{\theta} > \frac{-2R_{tol}}{||\boldsymbol{X}_n - \boldsymbol{X}_o||_2} v_r \text{ or } |\gamma| > \frac{5\pi}{8}\\
        f K_{vor}^{0}  \quad \text{ otherwise} 
    \end{cases}
    \label{eq:modified_vortex}
\end{align}
where $f$ is a factor to scale up the vortex strength depending on the collision risk and it is given by \eqref{eq:factor_f}.
\begin{align}
    f = \max\left\{1, \left(2 - \frac{||\boldsymbol{X}_n - \boldsymbol{X}_o||_2}{R_{safe}} - v_r\right)\right\}
    \label{eq:factor_f}
\end{align}

Note that $K_{vor}^{0} = -10$ is used in this study. The factor $f$ linearly increases as the distance between the vessel and obstacle decreases. Similarly, the factor $f$ increases as the radial component of the relative velocity becomes more and more negative. Note that the threshold $\frac{-2R_{tol}}{||\boldsymbol{X}_n - \boldsymbol{X}_o||_2} v_r$ on $v_{\theta}$ is chosen to ensure that the vessel is able to move out of the collision radius $R_{tol}$ within the time taken to cover the radial distance. The factor of $2$ ensures that the margin for safety in the tangential direction is twice the collision threshold. 

It is also important to note that in addition to the constraint on tangential velocity, the vortex potential is considered only when $|\gamma| \leq 5\pi/8$, which corresponds to either head on or crossing encounters. When $|\gamma| > 5\pi/8$, the encounter is considered to be an overtaking scenario as per COLREGS and hence the vessel has a stand on responsibility. This is satisfied by specifying the vortex strength as zero when $|\gamma| > 5\pi/8$.

\begin{figure}
    \centering
    \includegraphics[width=\linewidth]{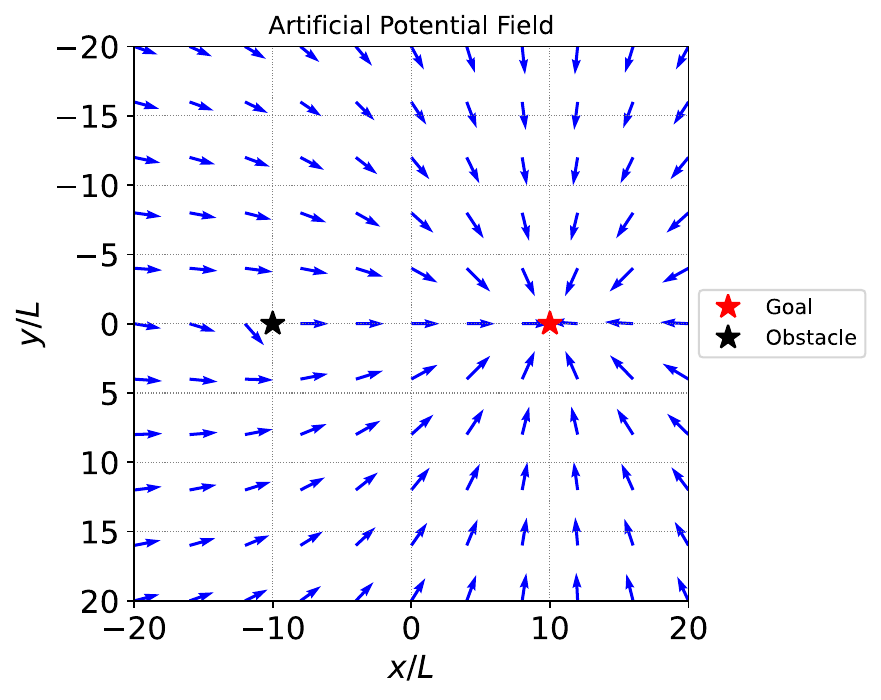}
    \caption{Vector field due to sink at goal and modified vortex center at obstacle}
    \label{fig:streamlines_mvortex}
\end{figure} 

\begin{figure}
    \centering
    \includegraphics[width=\linewidth]{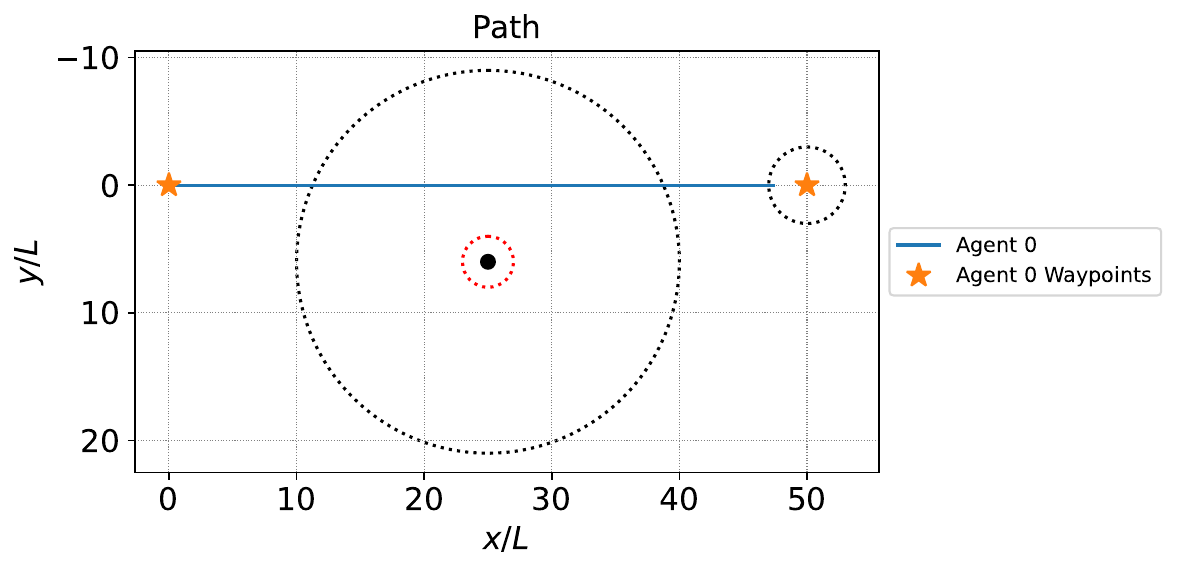}
    \caption{Modified sink-vortex APF method tracks the waypoint in the case where traditional sink-vortex APF method fails}
    \label{fig:mvortex_static_success}
\end{figure}

The vector field for this formulation of modified vortex strength $K_{vor}$ is shown in \autoref{fig:streamlines_mvortex}. It can be seen that the vector field gradients do not point towards the obstacle when the obstacle is on the starboard side of the vessel. The modified vortex method is applied to the case where the vortex method leads to a collision and the results are shown in \autoref{fig:mvortex_static_success}. It can be seen that in this case the tangential relative velocity will be positive during the encounter and hence the vessel follows a straight line to the goal. 

\begin{figure*}
    \centering
    \begin{subfigure}{\textwidth}
      \centering
      \includegraphics[width=\textwidth]{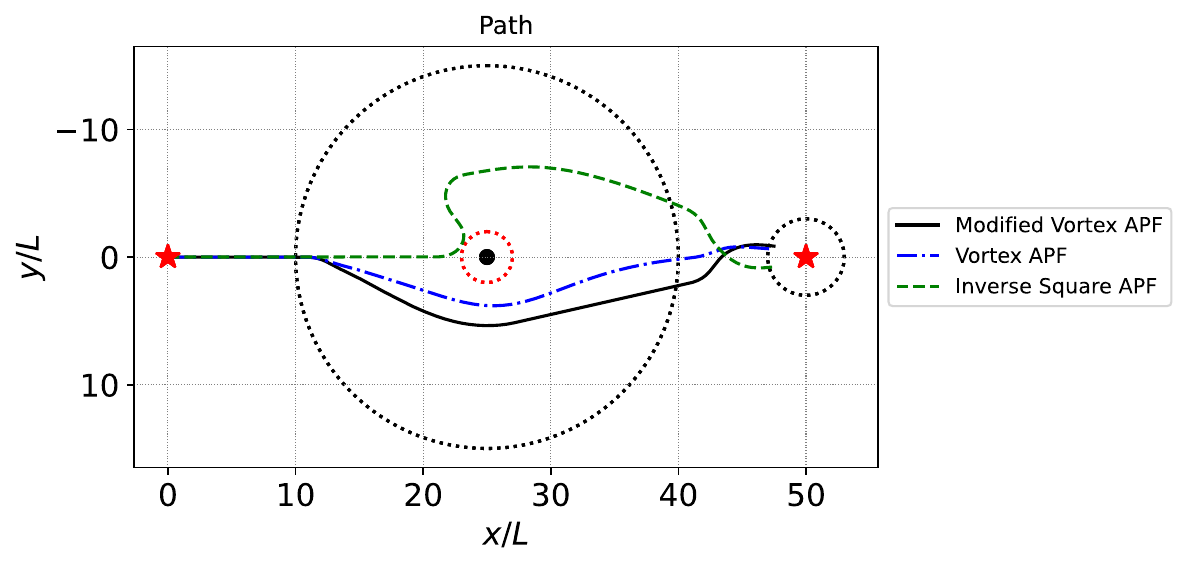}
      \caption{Vessel Trajectory: Evaluating the impact of different proposed potentials on the vessel trajectory.}
      \label{fig:harmonic_trajectory}
    \end{subfigure}
    \vspace{0.7cm}
    
    \begin{subfigure}{0.49\textwidth}
      \centering
      \includegraphics[width=\linewidth]{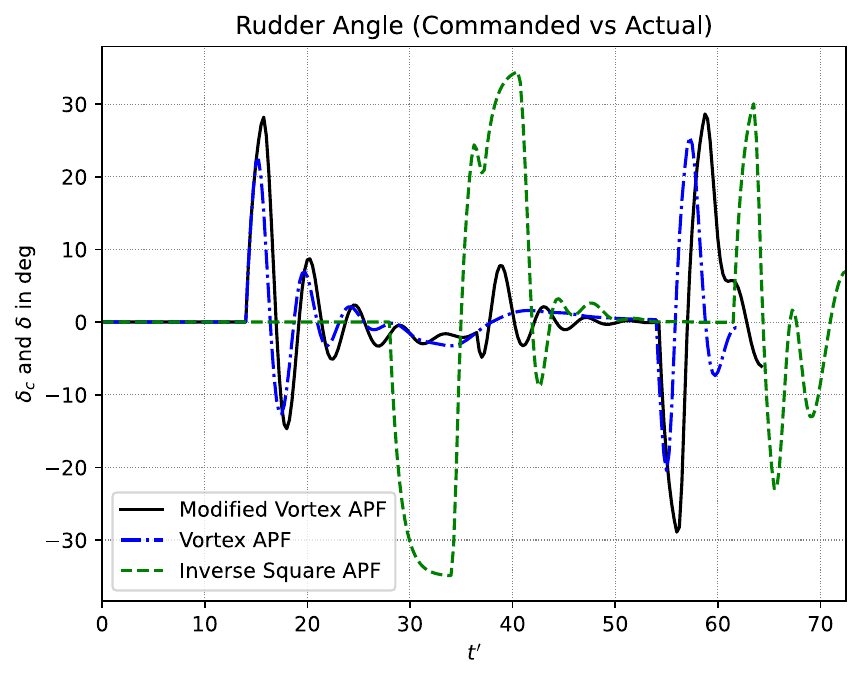}
      \caption{Applied rudder angle}
      \label{fig:harmonic_rudder}
    \end{subfigure}
    \hfill
    \begin{subfigure}{0.49\textwidth}
      \centering
      \includegraphics[width=\linewidth]{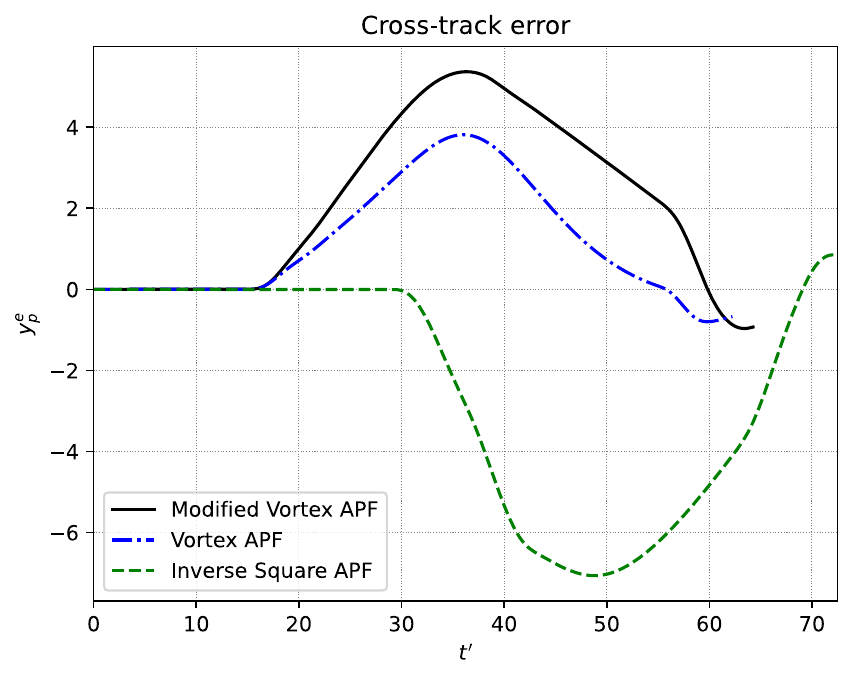}
      \caption{Cross-track Error Variation}
      \label{fig:harmonic_ype}
    \end{subfigure}
    
    \caption{Static Obstacle Avoidance: A comparison of vessel performance using the three APFs - inverse square, sink-vortex and modified sink-vortex}
    \label{fig:harmonic_main}
\end{figure*}

Using the dynamics described in \eqref{eq:mmg_nd} and the theory described above, the vessel is simulated in a static obstacle-based environment using three potential fields - inverse square, sink-vortex and modified sink-vortex. The environment setup is kept the same as explained in \autoref{fig:inverse_square_main} such that an obstacle encounter was forced.
The resultant trajectories and their comparison can be seen in \autoref{fig:harmonic_main}.
The applied rudder and cross-track error variation are also compared.
The APF parameters used can be seen in \autoref{tab:static_obstacles_parameters} with a safe radius of $R_{safe} = 15 \text{ ship lengths}$.
It can be seen that the modified sink-vortex approach is a balance between the inverse square method and the traditional sink-vortex approach. The modified approach has a greater safety than the traditional vortex due to the additional factor $f$ defined in \eqref{eq:factor_f}.

\begin{table}[htbp]
    \renewcommand{\arraystretch}{1.2}
    \caption{APF parameters used for Static Obstacle Avoidance}
    \centering
    \begin{tabular*}{\tblwidth}{@{} LLLL@{} }
    \toprule
    \textbf{Environment} & \textbf{Parameters}  \\
    \midrule
    \multirow{2}{15em}{Sink + Vortex}  & $K_{vor} = -10$ \\
     & $\Lambda_{\text{sink}} = -100$ \\
    \midrule
    \multirow{2}{15em}{Sink + Modified Vortex}  & $K_{vor} = -10$ \\
     & $\Lambda_{\text{sink}} = -100$ \\
    \midrule
    \multirow{2}{15em}{Inverse Square} & $k_{\text{att}} = 50$  \\
     & $k_{\text{rep}} = 200000$ \\
    \midrule
   \end{tabular*}
    \label{tab:static_obstacles_parameters}
\end{table}

\section{Dynamic Obstacle Avoidance}
\label{sec:dynamic_obs}
This section discusses additional ability of the earlier described APFs to consider dynamic obstacles present in the environment. The potential calculation for the static obstacles follows from the theory described in \Secref{sec:Harmonic Functions}.

The dynamic obstacle avoidance capability using the above-proposed method is tested on the three rules specified by the International Maritime Organization (IMO) in the Convention on the International Regulations for Preventing Collisions at Sea (COLREGs):
\begin{enumerate}
    \item \textbf{Rule 13 - Overtaking:}  \textit{Any vessel overtaking any other shall keep out of the way of the vessel being overtaken}
    \item \textbf{Rule 14 - Head on Situation:} \textit{When two power-driven vessels are meeting on reciprocal or nearly reciprocal courses so as to involve risk of collision each shall alter her course to starboard so that each shall pass on the port side of the other}
    \item \textbf{Rule 15 - Crossing Situation:} \textit{When two power-driven vessels are crossing so as to involve risk of collision, the vessel which has the other on her own starboard side shall keep out of the way and shall, if the circumstances of the case admit, avoid crossing ahead of the other vessel}
\end{enumerate}

\begin{figure*}[htbp]
    \centering
    \begin{subfigure}{0.9\textwidth}
        \centering
        \includegraphics[width=\textwidth]{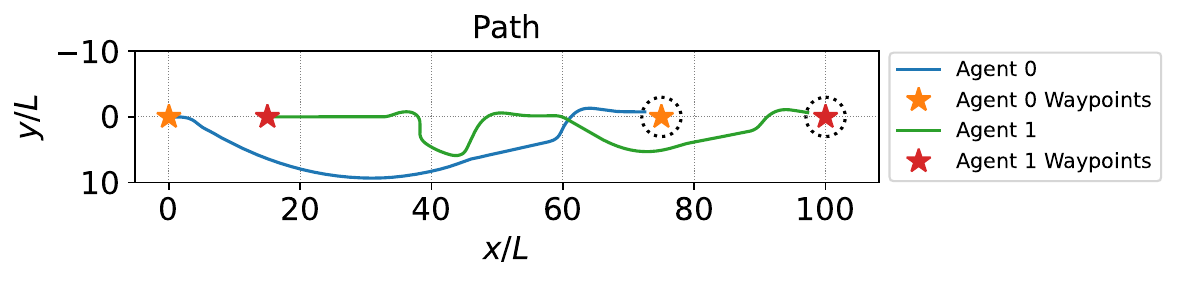}
        \caption{Overtaking Situation}
    \end{subfigure}
    \begin{subfigure}{0.9\textwidth}
        \centering
        \includegraphics[width=\textwidth]{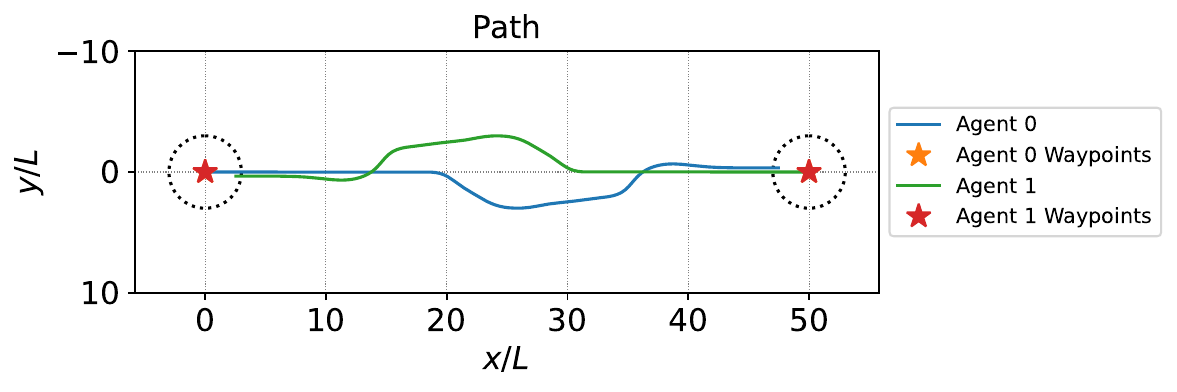}
        \caption{Head On Situation}
    \end{subfigure}
    \begin{subfigure}{0.7\textwidth}
        \centering
        \includegraphics[width=\textwidth]{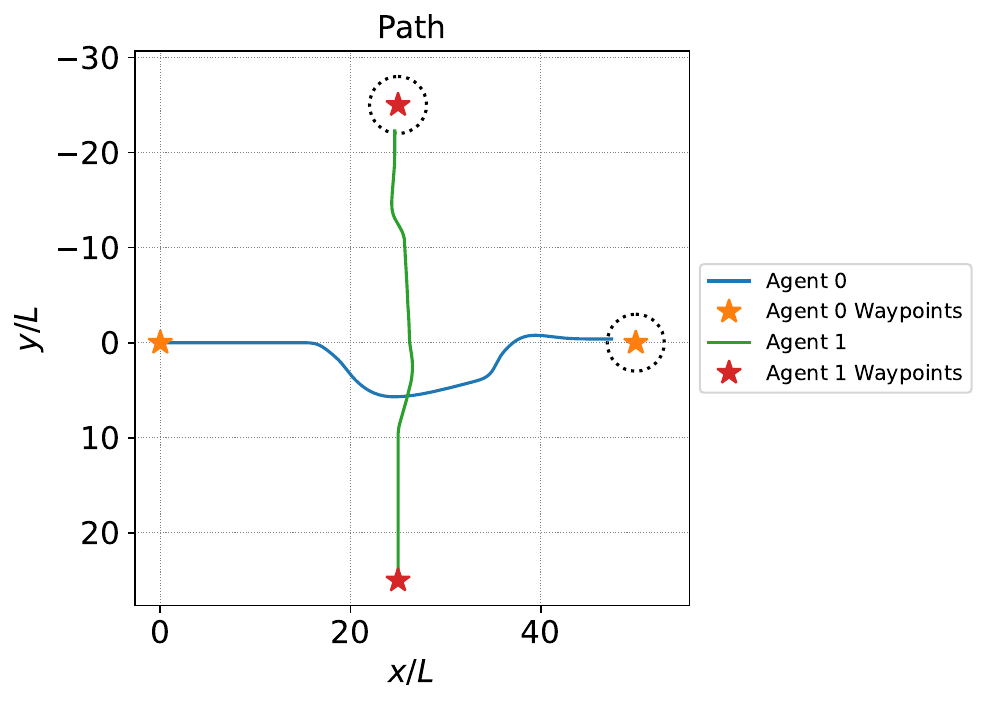}
        \caption{Crossing Encounter}
    \end{subfigure}    
    \caption{Path taken during dynamic obstacle avoidance using modified sink-vortex APF in overtaking, head on and crossing situations}
    \label{fig:mvortex_colregs_path}
\end{figure*}

\begin{figure*}
    \centering
    \begin{subfigure}{0.48\textwidth}
        \centering
        \includegraphics[width=\textwidth]{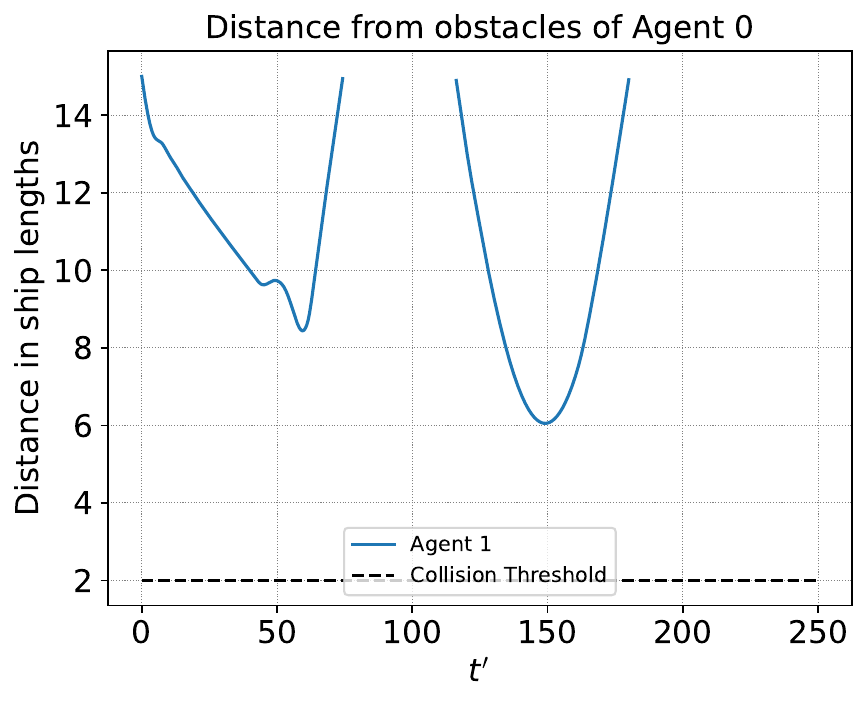}
        \caption{Overtaking Situation}
    \end{subfigure}
    \begin{subfigure}{0.48\textwidth}
        \centering
        \includegraphics[width=\textwidth]{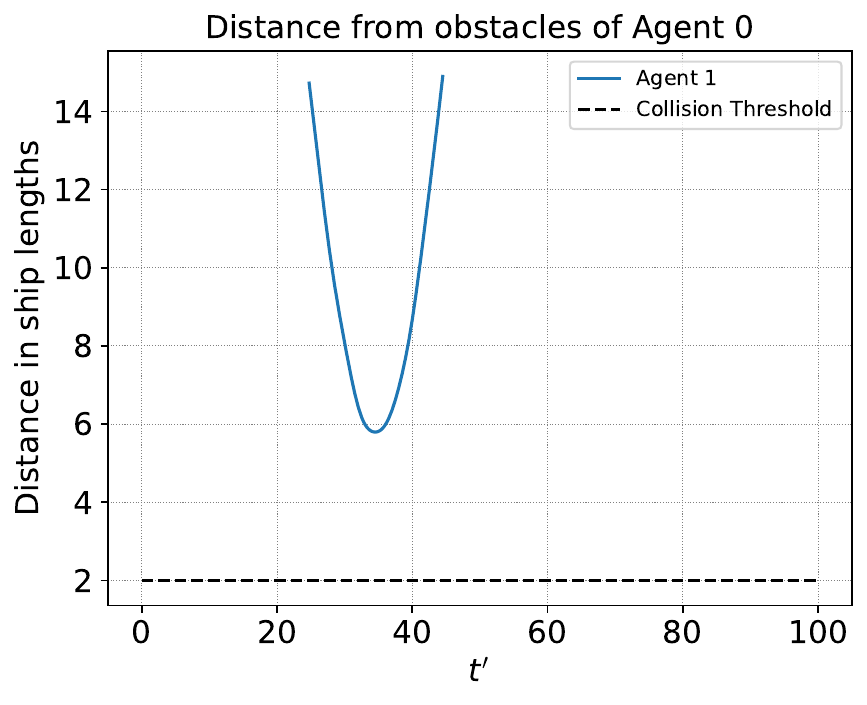}
        \caption{Head On Situation}
    \end{subfigure}
    \begin{subfigure}{0.48\textwidth}
        \centering
        \includegraphics[width=\textwidth]{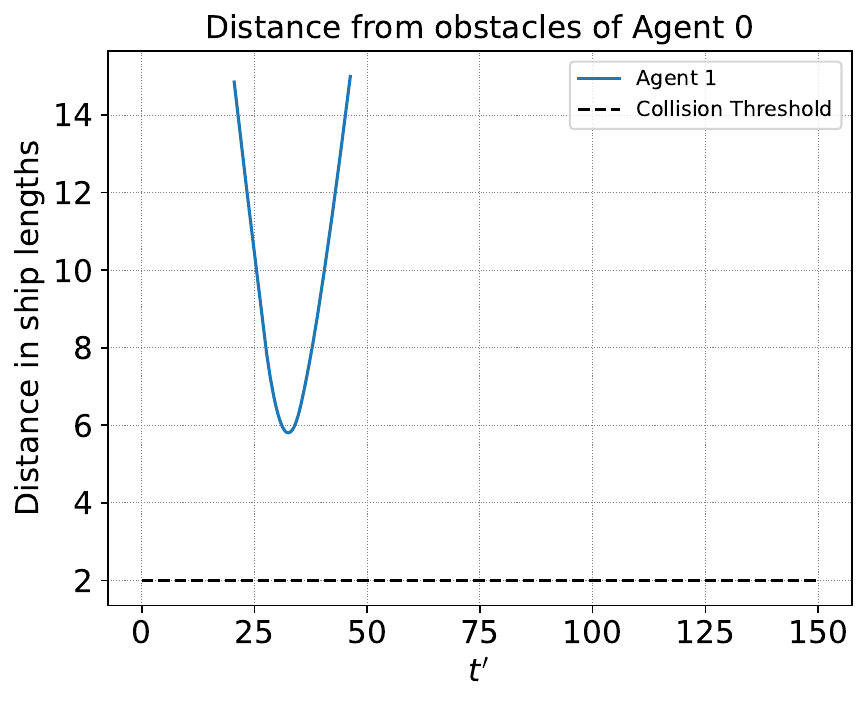}
        \caption{Crossing Encounter}
    \end{subfigure}    
    \caption{Distance between agents during dynamic obstacle avoidance (only while within $R_{safe}=15L$ distance of each other) using modified sink-vortex APF in overtaking, head on and crossing situations}
    \label{fig:mvortex_colregs_dist}
\end{figure*}

The paths taken by modified sink-vortex APF in overtaking, head on and crossing situations are shown in \autoref{fig:mvortex_colregs_path}. The corresponding distance between agents in each of these encounters are shown in \autoref{fig:mvortex_colregs_dist}. The APF parameters used in the simulation are the same as those used in static obstacle avoidance, as seen in \autoref{tab:static_obstacles_parameters}.

In the overtaking scenario, the overtaking vessels tries to cross the overtaken vessel on the starboard side of the overtaken vessel. Both vessels in this case are assumed to be the KCS vessel with the overtaken vessel having a operating speed of half the design speed of the overtaking vessel. The minimum distance between the vessels in this encounter is little more than $8$ ship lengths. 

In the head on encounter, both vessels have the same speed (equal to the design speed). The first vessel starts at $(0L, 0L)$ and tracks its goal waypoint $(50L, 0L)$. For the second vessel the start location is $(50L, 0L)$ and the goal waypoint is $(0L,0L)$. Thus, without a collision avoidance mechanism the vessels would collide at $(25L, 0L)$. It can be seen that with modified sink-vortex APF, both vessels take a starboard turn when they are $10$ ship lengths away from each other. The minimum distance between the vessels is approximately $6$ ship lengths.

In the crossing encounter, the first vessel starts at $(0L, 0L)$ with a goal of $(50L, 0L)$ while the second vessel starts at $(25L, 25L)$ with a goal of $(25L, -25L)$. Without a collision avoidance mechanism the vessels would collide at $(25L, 0L)$. Under the influence of the modified sink-vortex APF, the first vessel takes a starboard turn to avoid the second vessel that is coming from the starboard side of the first vessel. The second vessel, however, does not deviate significantly from its course. The minimum distance between the vessels in this case too is approximately $6$ ship lengths. 

\begin{figure*}
    \centering
    \begin{subfigure}{0.9\textwidth}
        \centering
        \includegraphics[width=\textwidth]{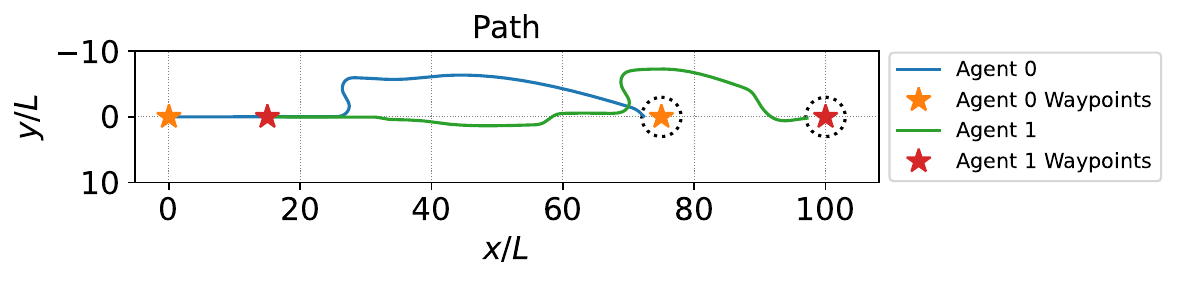}
        \caption{Overtaking Situation}
    \end{subfigure}
    \begin{subfigure}{0.9\textwidth}
        \centering
        \includegraphics[width=\textwidth]{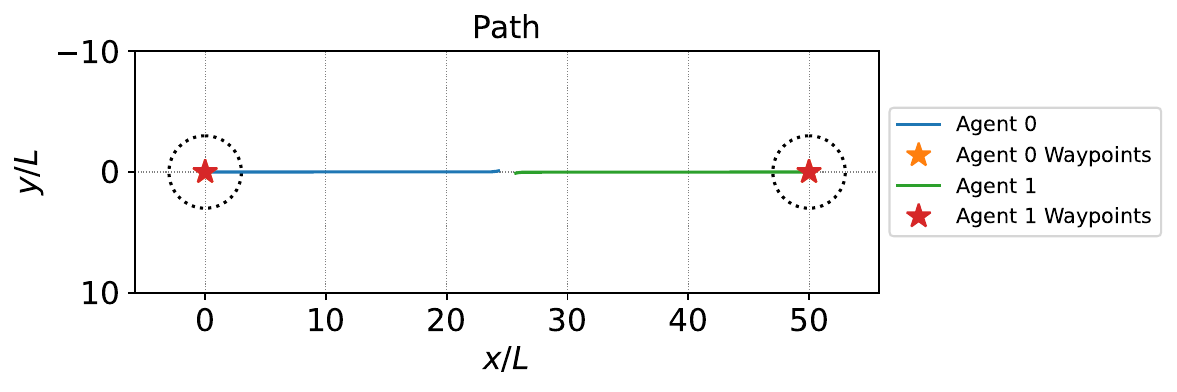}
        \caption{Head On Situation}
    \end{subfigure}
    \begin{subfigure}{0.65\textwidth}
        \centering
        \includegraphics[width=\textwidth]{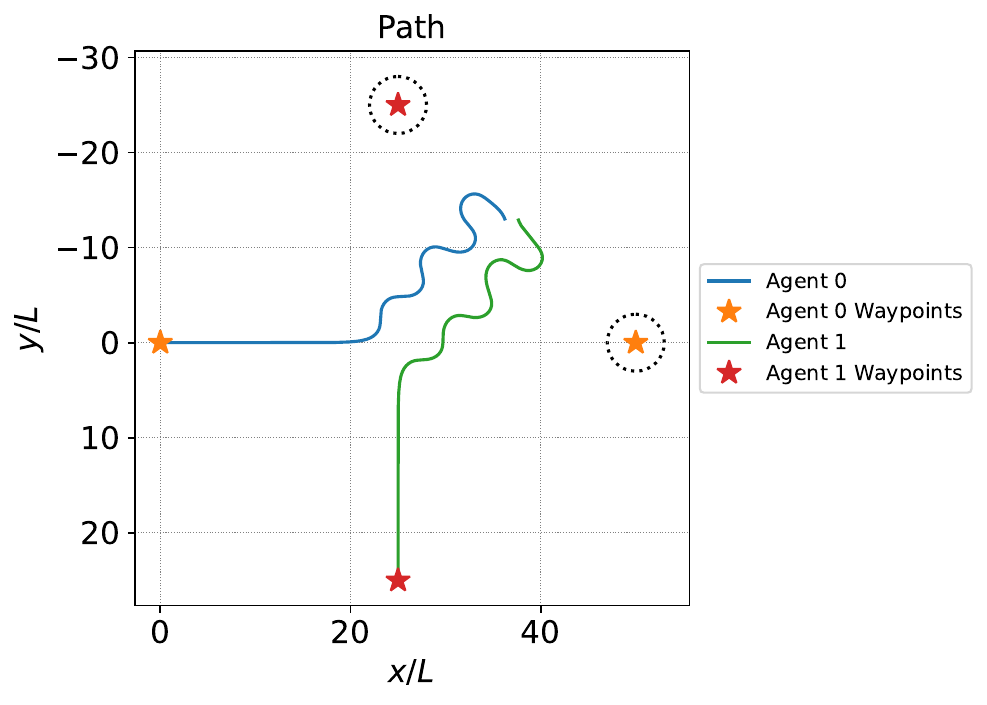}
        \caption{Crossing Encounter}
    \end{subfigure}    
    \caption{Path taken during dynamic obstacle avoidance using inverse square APF in overtaking, head on and crossing situations}
    \label{fig:inverse_square_colregs_path}
\end{figure*}

In order to compare its performance against the traditional APF, the same encounters are also tested with the inverse square APF. The start waypoints, goal waypoints and the vessel initial conditions are the same as chosen in the modified sink-vortex case. The paths followed are shown in \autoref{fig:inverse_square_colregs_path}. Note that the APF parameters used in the simulation are the same as those used in static obstacle avoidance and are reported in \autoref{tab:static_obstacles_parameters}.

Unlike the modified sink-vortex APF case, here the overtaking vessel tries to cross from the port side of the overtaken vessel. The minimum distance between the vessels in this encounter is approximately $3$ ship lengths. In both the head on collision and in the crossing encounter, a collision occurs as the distance between the ships reduces to less than $2$ ship lengths. In the head on case, the vessels do not execute a turn until it is too late as the gradients due to the attractive potential driving the vessel towards the goal dominate the repulsive gradients due to the obstacles. Note that this is a degenerate case where all the gradients are oriented along the x-axis. This causes the decision to change course to be initiated only when the vessels are too close. However, due to the significant inertia, the vessels are unable to change course before the distance between them reduces below the tolerance for collision. In the crossing encounter, the vessels do experience the effect of the repulsive potential as the direction of the repulsive gradient is different from the direction of the attractive gradient and this causes the vessels to approach and go away from each other multiple times before the distance between them reduces below the threshold of $2$ ship lengths.

In addition to the traditional inverse square APF, the same encounters are also simulated using the velocity obstacle method to compare the modified sink-vortex APF method against another reactive guidance approach. The modelling details of the velocity obstacle method follow the linear velocity obstacle (LVO) approach as described by \cite{huang2018velocity} and are avoided here for brevity. Since all simulations for APF simulations are performed under the assumption of a constant RPM, the same assumption is maintained for the velocity obstacle approach too. This is achieved through an additional constraint on the magnitude of the velocity vector. This favours picking an acceptable velocity vector through course change rather than speed change.

\begin{figure*}
    \centering
    \begin{subfigure}{0.9\textwidth}
        \centering
        \includegraphics[width=\textwidth]{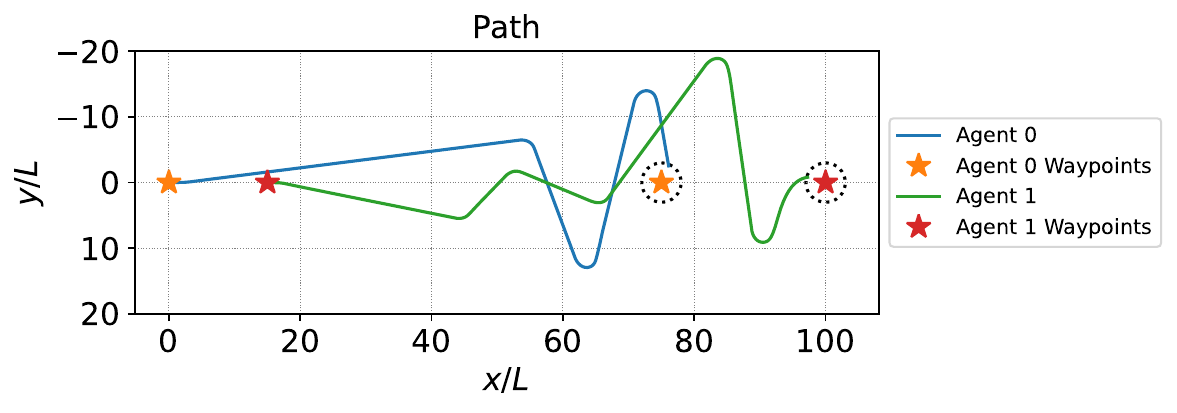}
        \caption{Overtaking Situation}
    \end{subfigure}
    \begin{subfigure}{0.9\textwidth}
        \centering
        \includegraphics[width=\textwidth]{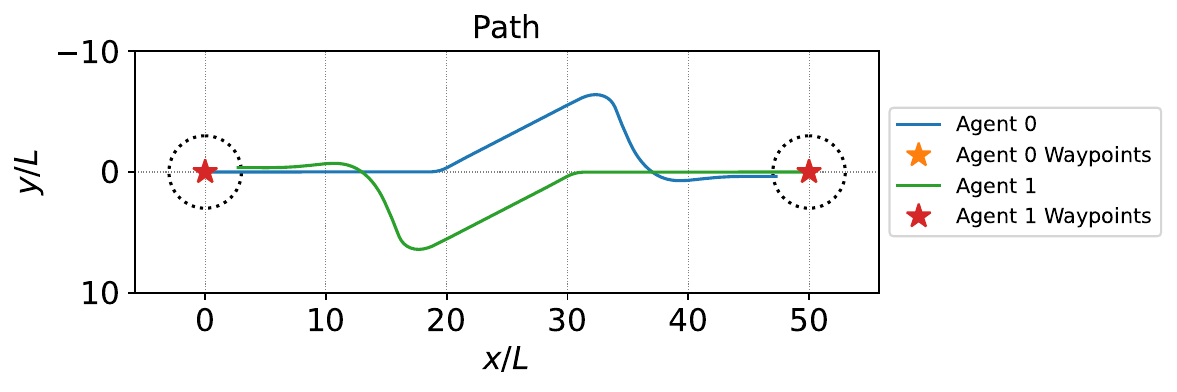}
        \caption{Head On Situation}
    \end{subfigure}
    \begin{subfigure}{0.65\textwidth}
        \centering
        \includegraphics[width=\textwidth]{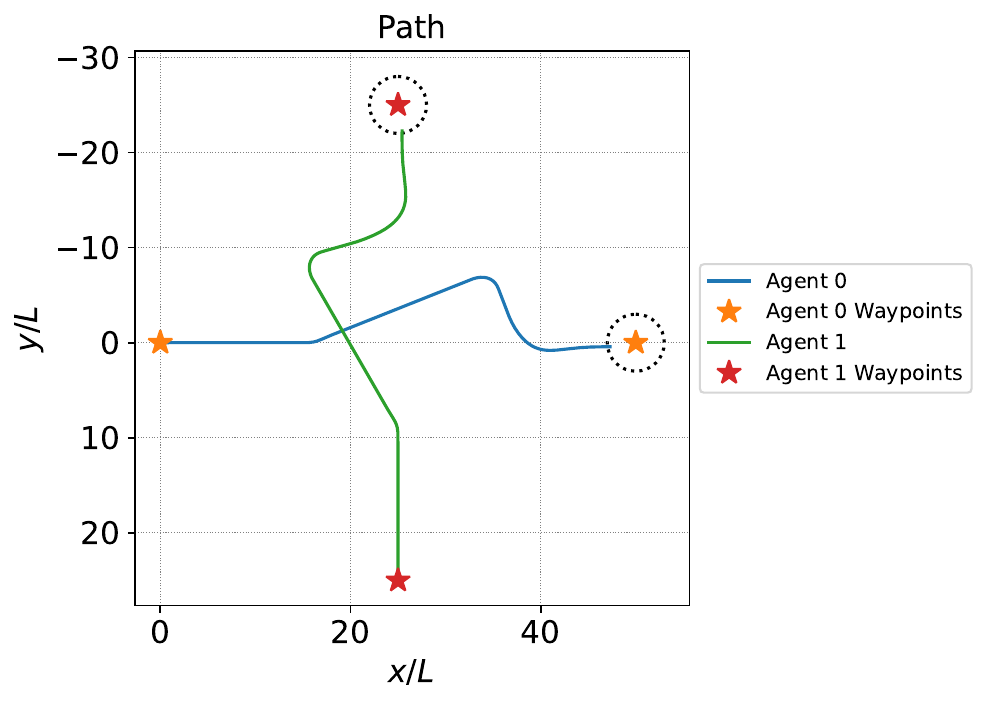}
        \caption{Crossing Encounter}
    \end{subfigure}    
    \caption{Path taken during dynamic obstacle avoidance using velocity obstacle in overtaking, head on and crossing situations}
    \label{fig:velocity_obstacle_colregs_path}
\end{figure*}

\begin{figure*}
    \centering
    \begin{subfigure}{0.48\textwidth}
        \centering
        \includegraphics[width=\textwidth]{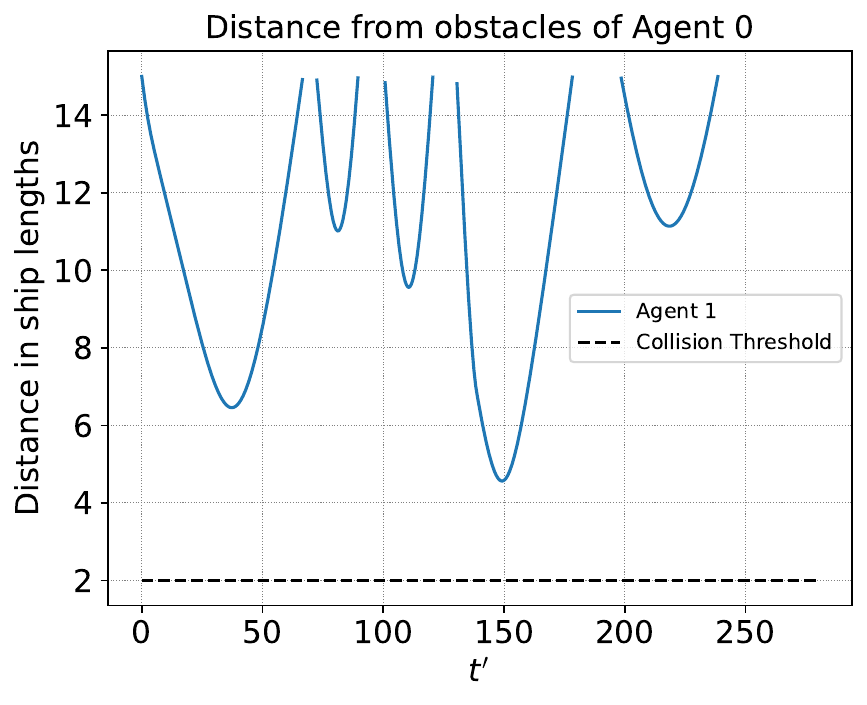}
        \caption{Overtaking Situation}
    \end{subfigure}
    \begin{subfigure}{0.48\textwidth}
        \centering
        \includegraphics[width=\textwidth]{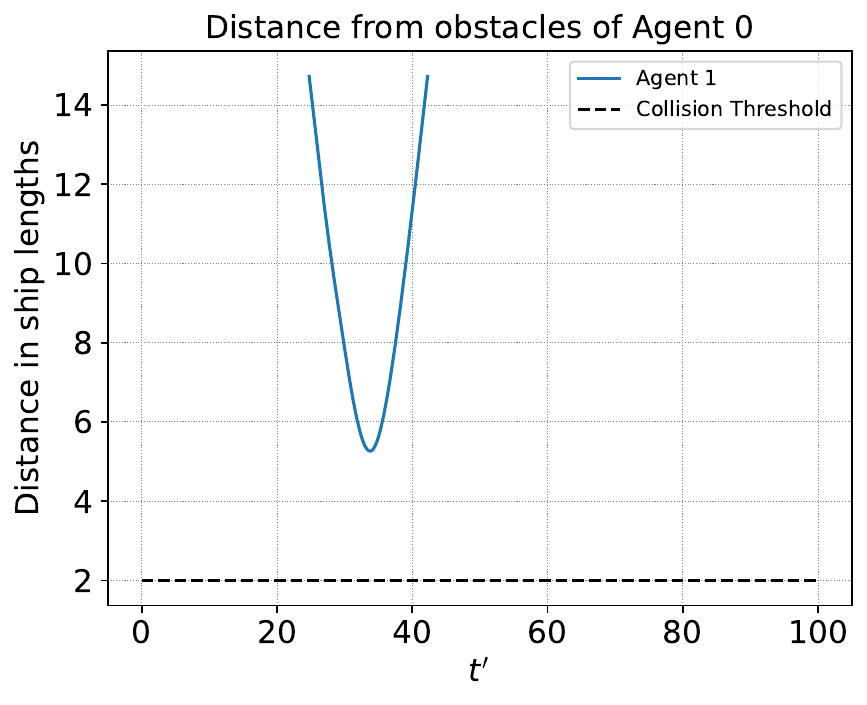}
        \caption{Head On Situation}
    \end{subfigure}
    \begin{subfigure}{0.48\textwidth}
        \centering
        \includegraphics[width=\textwidth]{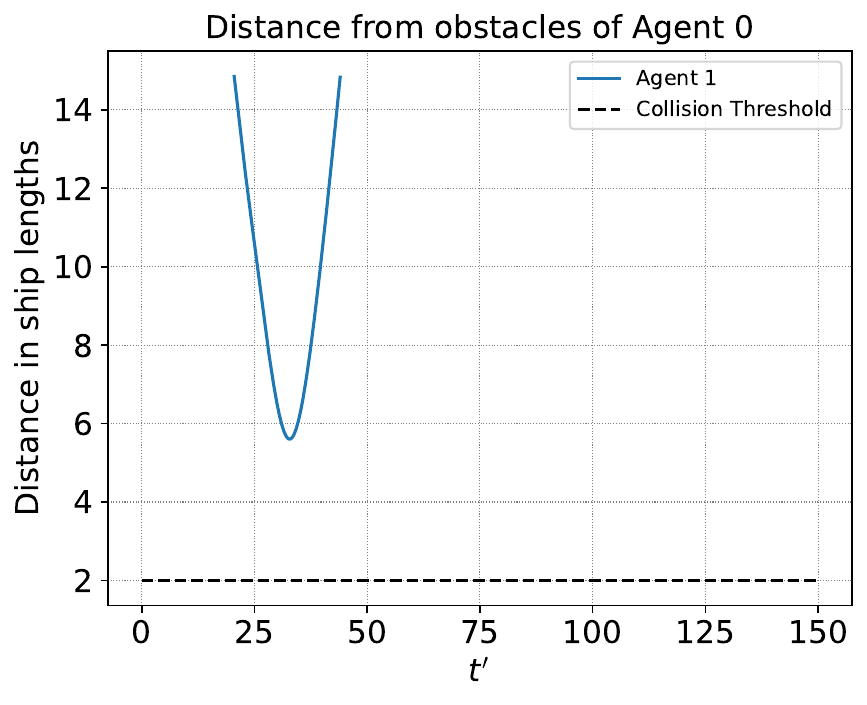}
        \caption{Crossing Encounter}
    \end{subfigure}    
    \caption{Distance between agents during dynamic obstacle avoidance (only while within $R_{safe}=15L$ distance of each other) using velocity obstacle method in overtaking, head on and crossing situations}
    \label{fig:velocity_obstacle_colregs_dist}
\end{figure*}

In the overtaking case, it is observed that both the vessels change heading multiple times to avoid collision with each other until the first ship (overtaking vessel) tracks its goal. After the first ship has tracked its goal, the second vessel (overtaken vessel) proceeds to its original heading after moving out of the safe radius $R_{safe}=15L$ of the first ship. This encounter clearly does not follow COLREG rules as the overtaken ship has a stand on responsibility and should not take any action in this scenario. The minimum distance between the agents is found to be $4L$ as seen from \autoref{fig:velocity_obstacle_colregs_dist}. This limitation is a well understood behaviour in velocity obstacle method \citep{huang2018velocity}.

In the head on encounter, both the vessels change course towards their port side until they are clear of each other and return to their original course afterwards. The minimum distance between the vessels is about $6L$ and is similar to as observed from the modified sink-vortex APF method. In the crossing encounter, both the vessels take a port turn to avoid collision with each other. In this scenario the minimum distance between the vessels is observed to be $6L$ and is similar to as observed with the modified sink-vortex APF method.

Thus, the modified sink-vortex APF is able to successfully follow COLREG rules 13, 14 and 15. However, this is not the case with inverse square APF. The velocity obstacle method also is able to achieve a collision free path, but does not necessarily follow COLREG rules. 


\subsection{Conflicting Responsibility}

COLREGS specifies a generic set of rules for all vessels to follow. However, at times vessels may encounter conflicting responsibility when multiple vessels might be interacting. In order to investigate this effect, a sample scenario involving three ships is constructed that results in a conflict of responsibility. The first ship starts at $(0,0)$ and tracks the waypoint $(60L, 0)$. The second ship which is moving at twice the speed of the first ship starts at $(-20L, 0)$ is tracking the waypoint $(100L, 0)$. The third ship moving at twice the speed of the first ship starts at $(60L, 0)$ and tracks the waypoint $(0,0)$. Without a collision avoidance strategy all three ships would collide with each other at $(20L, 0)$. Thus the first vessel has stand on responsibility with respect to second ship while it has give way responsibility with respect to third ship. Note that it is assumed that all three vessels have the same length.

\begin{figure*}
    \centering
    \begin{subfigure}{0.9\textwidth}
        \centering
        \includegraphics[width=\textwidth]{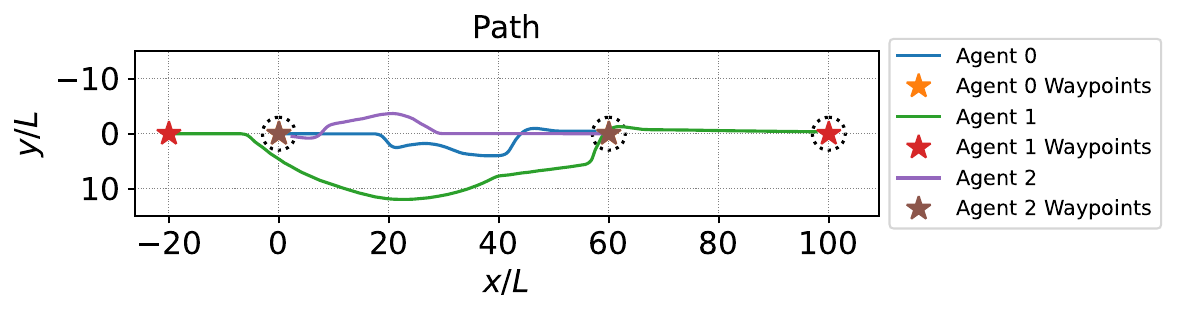}
        \caption{Modified sink-vortex APF}
    \end{subfigure}
    \begin{subfigure}{0.9\textwidth}
        \centering
        \includegraphics[width=\textwidth]{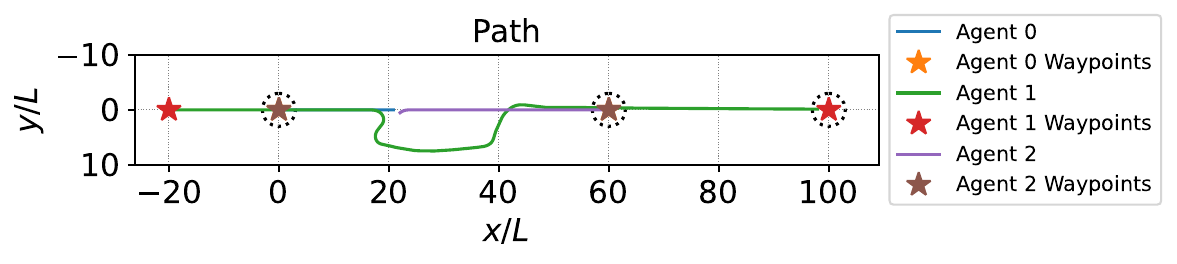}
        \caption{Traditional inverse square APF}
    \end{subfigure}
    \begin{subfigure}{0.9\textwidth}
        \centering
        \includegraphics[width=\textwidth]{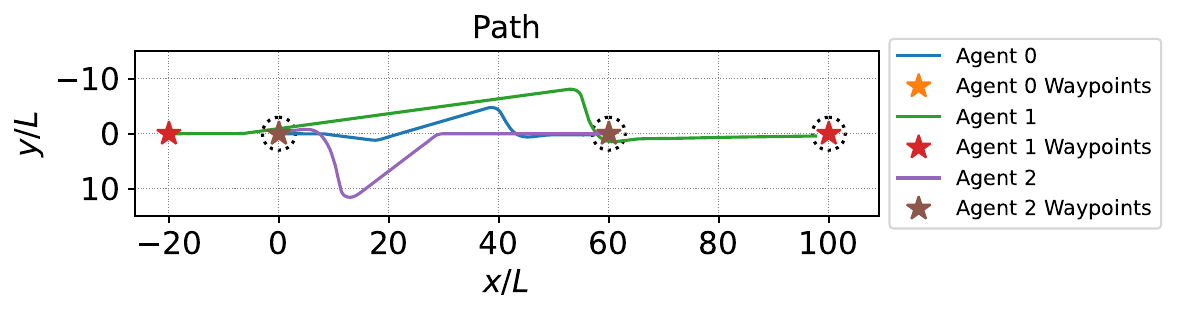}
        \caption{Velocity Obstacle}
    \end{subfigure}
    \caption{Conflicting responsibility in an encounter}
    \label{fig:colregs_conflict}
\end{figure*}

The same encounter is investigated using traditional APF, modified sink-vortex APF and the velocity obstacle method. A comparison of the paths between the three approaches is shown in \autoref{fig:colregs_conflict}.

It can be seen that with traditional APF, the first and the third ships collide while the second ship is able to avoid collision in the nick of the time. The velocity obstacle method results in collision free paths for all vessels. However, it is seen that the approach does not follow the COLREG rules. The modified sink-vortex APF method results in a collision free path for all the vessels while following COLREG rules. 

\begin{figure*}
    \centering
    \begin{subfigure}{0.45\textwidth}
        \centering
        \includegraphics[width=\textwidth]{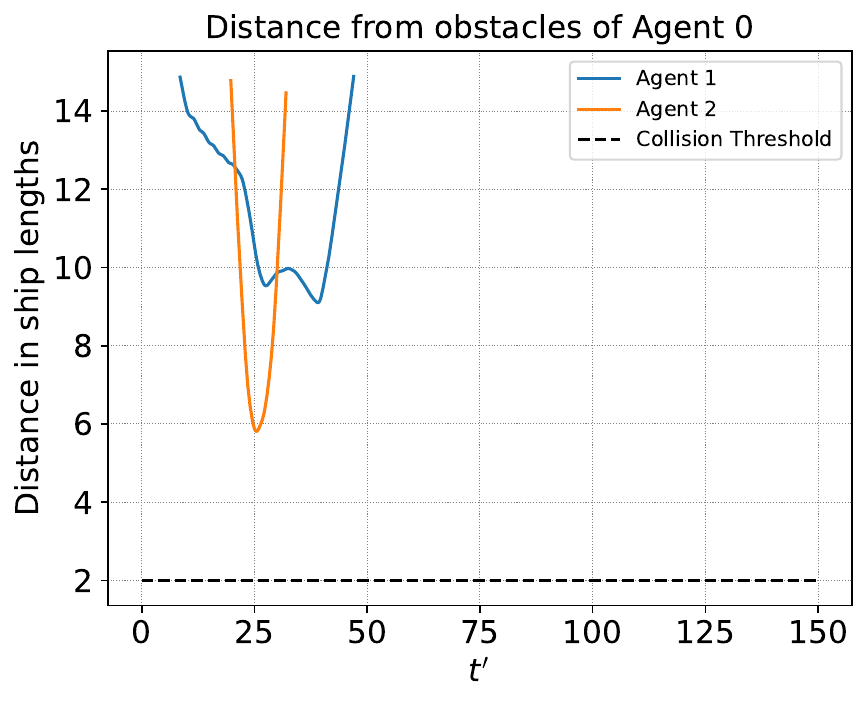}
        \caption{Modified sink-vortex APF}
    \end{subfigure}
    \begin{subfigure}{0.45\textwidth}
        \centering
        \includegraphics[width=\textwidth]{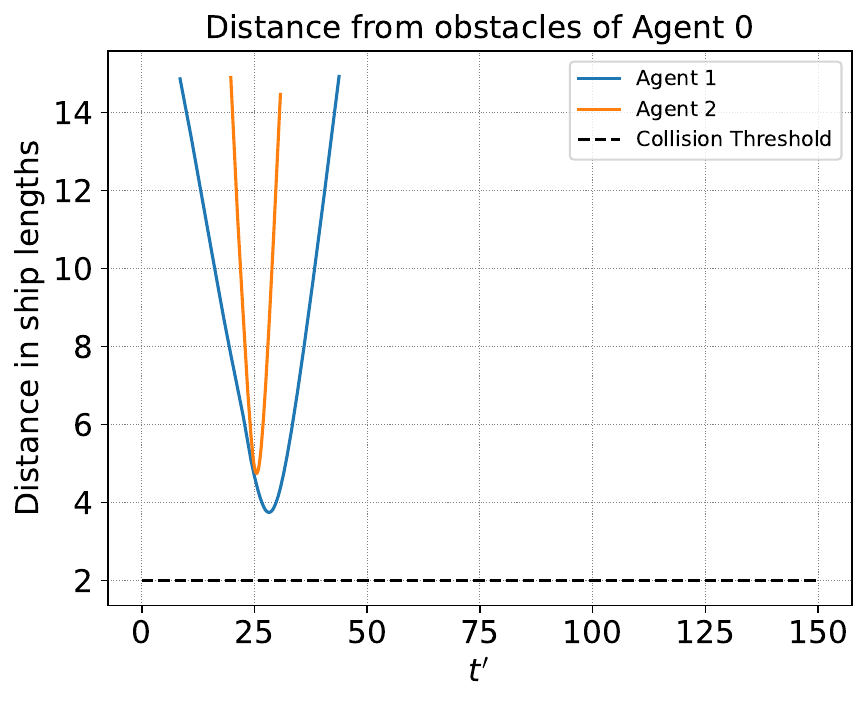}
        \caption{Velocity Obstacle}
    \end{subfigure}
    \caption{Distance between the vessels while inside the safe radius in an encounter that has conflicting responsibility as shown in \autoref{fig:colregs_conflict}}
    \label{fig:colregs_conflict_dist}
\end{figure*}

The distance between first ship and the other two vessels for modified sink-vortex APF method and velocity obstacle method are shown in \autoref{fig:colregs_conflict_dist}. It is seen that the minimum distance between the vessels is about $6L$ in case of modified sink-vortex APF method as compared to $4L$ in case of velocity obstacle method. There is a greater margin of safety between the ships in case of the modified sink-vortex APF method. It can be seen that the vortex potential causes the change in course to happen soon after the other vessel enters the safe radius and this leads to the margin of safety to be larger. 

\subsection{Narrow Waterways}

The discussion so far has assumed that there is sufficient space for maneuvering around the other vessel. However, during transit through narrow channels, there may not always be enough space to maneuver around the other vessel. In this section an example is demonstrated where two ships of same length $L$ cross each other in a head-on scenario in a narrow channel. The first ship starts at $(0,0)$ and tracks the waypoint at $(50L, 50L)$ and the second ship starts at $(50L, 50L)$ and tracks the waypoint at $(0,0)$. Both vessels start with an initial heading in the direction of the goal waypoint. It is assumed that the encounter occurs in a channel of width $10L$. 

The repulsive potential due to the channel boundaries are assumed to be similar to line sources being placed along the boundary. The repulsive potential is considered in the computation of gradients only when the vessel is within $2L$ distance from the boundary of the channel. At every point along the channel within this threshold distance of $2L$ from the boundary, the gradients are oriented perpendicular to the channel boundary and towards the center of the channel. The velocity gradient is still given by \eqref{eq:Vsink equation} where $r$ now represents the perpendicular distance of the point from the line source. The source strength for the line sources in this example is taken as $\Lambda = 10$.

\begin{figure*}
    \centering
    \begin{subfigure}{0.54\textwidth}
        \centering
        \includegraphics[width=\textwidth]{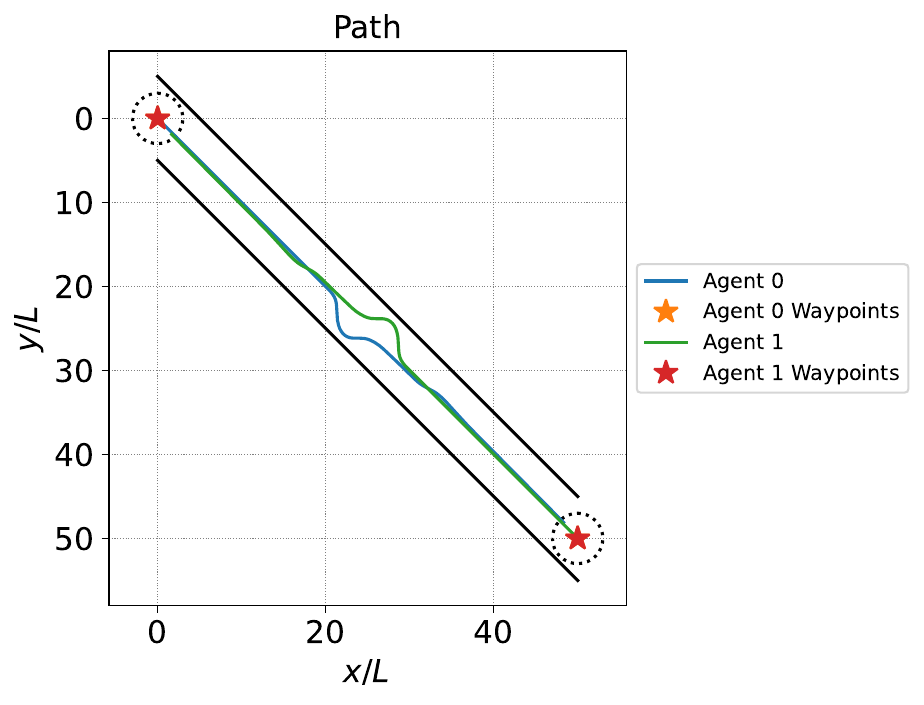}
        \caption{Encounter path in a narrow channel}
    \end{subfigure}
    \begin{subfigure}{0.45\textwidth}
        \centering
        \includegraphics[width=\textwidth]{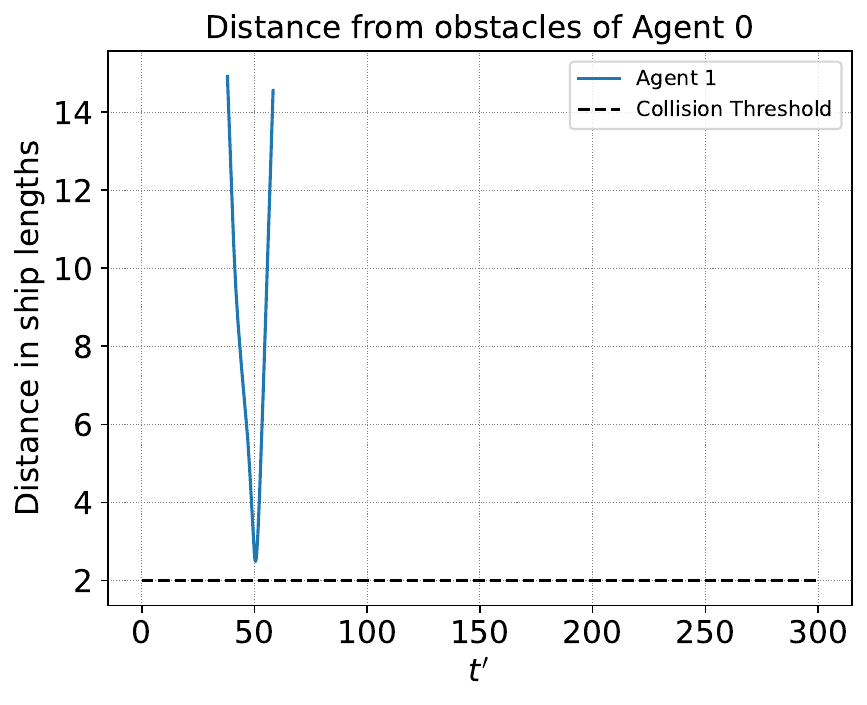}
        \caption{Distance between ships while inside safe radius during encounter}
    \end{subfigure}
    \caption{Head-on encounter in a narrow channel (Black lines represent the channel boundaries)}
    \label{fig:narrow_channel}
\end{figure*}

\autoref{fig:narrow_channel} shows the path of both vessels during the encounter and the distance between the vessels while they are within the safe radius $R_{safe}=15L$ of each other. It is seen that the vessels cross each other with a minimum distance of little over $2L$. This is in contrast to the minimum distance of $6L$ observed in open seas shown in \autoref{fig:mvortex_colregs_dist}. The effect of the boundary is clearly observed in the path taken by the vessels in \autoref{fig:narrow_channel} as compared to the encounter in open sea shown in \autoref{fig:mvortex_colregs_path}.

Although the APF method is shown to be successful in the narrow waterway, it is important to note that bathymetry effects have not been considered here. A changing bathymetry in the channel and the possibility of running aground will add further complexity to the problem. Although this can potentially be handled by specifying boundaries along isobaths, this is beyond the scope of this study. 

\section{Statistical Analysis}
\label{sec:statistical_analysis}
In order to test the effectiveness of the modified sink-vortex APF, a statistical analysis is performed over varied environments to compare it against the traditional APF method and the velocity obstacle method. Five different environments are considered with varying number of static obstacles and dynamic vessels in addition to the presence of the own ship. The number of static and dynamic obstacles (excluding the own ship) in each environment are listed in \autoref{tab:statistical_analysis_environment}. For each environment and a chosen method of collision avoidance, 5000 simulations are run. 

\begin{figure*}[!]
    \centering
    \includegraphics[width=0.7\textwidth]{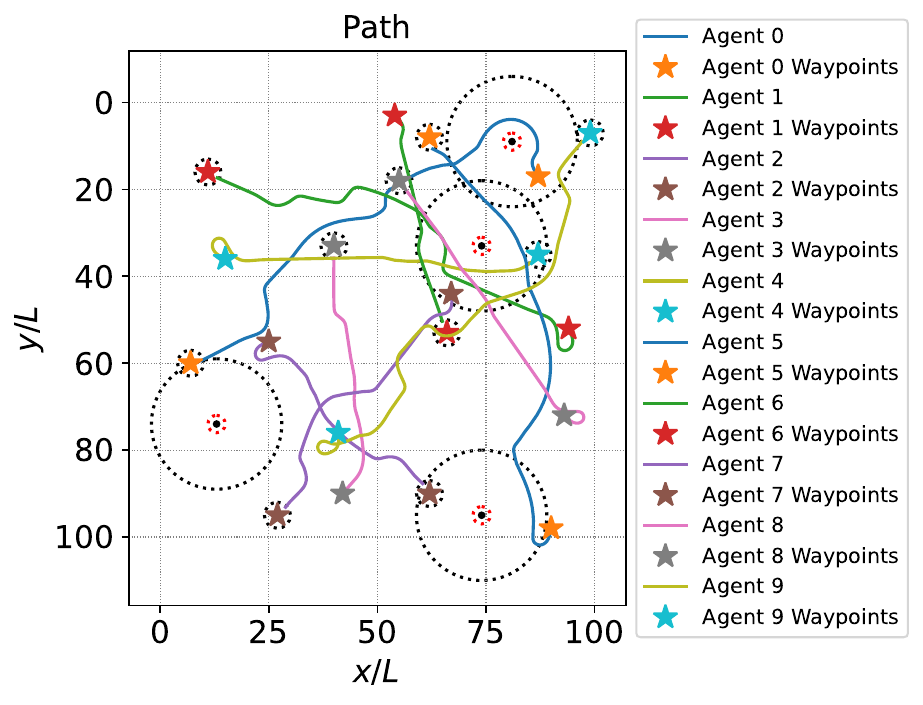}
    \caption{Sample case from statistical analysis demonstrating collision avoidance demonstrated using the modified sink-vortex APF architecture in Environment 5 with own ship (Agent 0), four static and nine dynamic obstacles (Agent 1 through 9). All initial waypoints are denoted by star markers while goal waypoints are denoted by star markers enclosed inside dotted circles. The static obstacles are denoted by black dots enclosed inside red dotted circles.}
    \label{fig:env_05_modified_vortex_result}
\end{figure*}

\begin{table}[htbp]
    \renewcommand{\arraystretch}{1.1}
    \caption{Environment Setup describing the number of static and dynamic obstacles in the vessel environment (excluding own ship)}    
    \begin{tabular*}{\tblwidth}{@{} LCC@{} }
    \toprule
    \textbf{Environment} & \textbf{Static Obstacles} & \textbf{Dynamic Obstacles} \\
    \midrule
        Environment 1 & 1 & 2 \\
        Environment 2 & 2 & 3 \\
        Environment 3 & 2 & 5 \\
        Environment 4 & 3 & 7 \\
        Environment 5 & 4 & 9 \\
    \bottomrule
   \end{tabular*}
    \label{tab:statistical_analysis_environment}
\end{table}
In each simulation the own ship, static obstacles and dynamic obstacles are spawned randomly in the environment. The initial positions of the own ship, static obstacles and dynamic obstacles are sampled from a uniform distribution across an area of $100L \times 100L$ where $L$ refers to the length of the own ship. When any of the distances between the initial states are less than $10L$ the points are re-sampled. This criterion is also enforced between starting states of dynamic vessels and static obstacles. This is needed to avoid scenarios where the vessels might be on unavoidable collision course at the start of the simulation. 

The radius of all static obstacles is assumed to be equal to $L/2$ in this study. The dynamic obstacles are assumed to have the same length as the own ship and are assigned a random speed sampled uniformly between $U_{des}/2$ and $U_{des}$ where $U_{des}$ is the design speed of the own ship. The goal points for the own ship and dynamic obstacles are also sampled uniformly across an area of $100L \times 100L$. If any of the goal waypoints are less than $50L$ from its corresponding initial waypoint, then the goal waypoints are re-sampled. The initial heading for own ship and dynamic obstacles are sampled from a uniform distribution ranging from $\left[-\pi, \pi \right]$.

All dynamic agents follow the same collision avoidance rules as the own ship. All the different parameters used for both the inverse square APF and the modified sink-vortex APF are same as listed in \autoref{tab:static_obstacles_parameters}. A sample result that uses modified sink-vortex APF method for collision avoidance in a random encounter scenario in Environment 5 can be seen in \autoref{fig:env_05_modified_vortex_result}, where the own ship navigates through four static obstacles and nine dynamic obstacles.

\begin{figure*}
    \centering
    \begin{subfigure}{0.9\textwidth}
      \centering
      \includegraphics[width=0.49\textwidth]{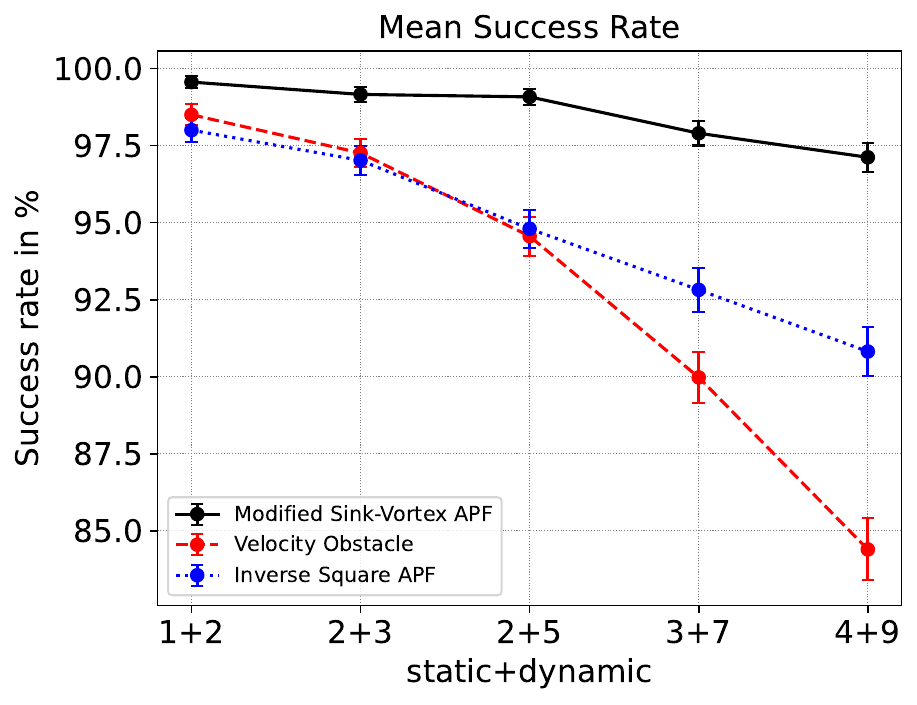}
      \caption{Success Rate}
      \label{fig:statistics_success_rate}
    \end{subfigure}
    \vfill
    \begin{subfigure}{0.45\textwidth}
      \centering
      \includegraphics[width=\textwidth]{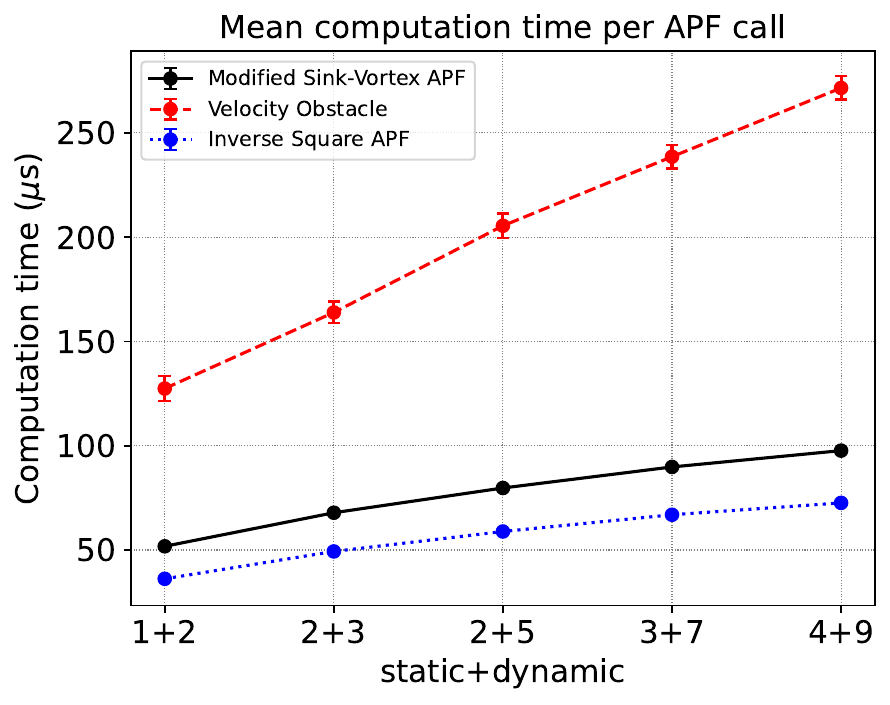}      
      \caption{Computation effort}
      \label{fig: statistics_computation}
    \end{subfigure}    
    \hfill    
    \begin{subfigure}{0.45\textwidth}
      \centering
      \includegraphics[width=\textwidth]{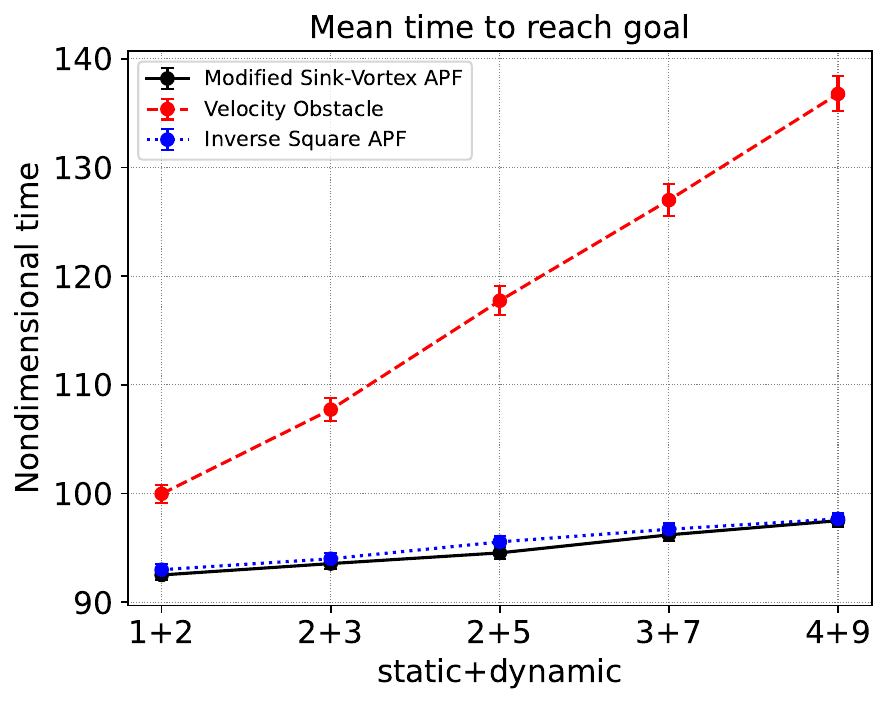}      
      \caption{Average non-dimensional time to reach the goal location}
      \label{fig:statistics_averaged_time_to_reach_goal}
    \end{subfigure}
    \vfill
    \begin{subfigure}{0.45\textwidth}
      \centering
      \includegraphics[width=\textwidth]{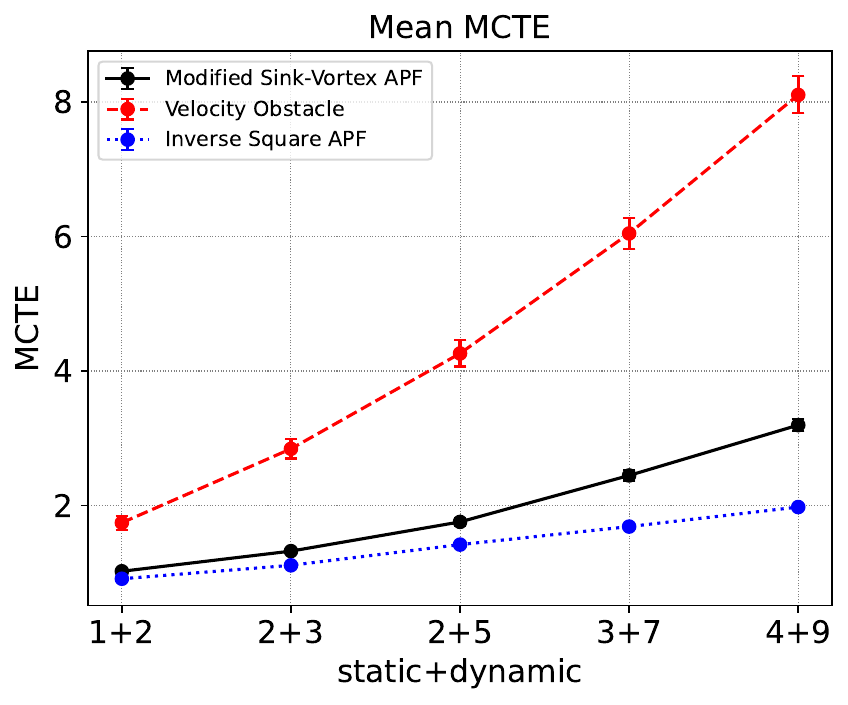}      
      \caption{Control Effort (CE) evaluated using \eqref{eq:controller_effort_equation}}
      \label{fig:statistics_control_effort}
    \end{subfigure}
    \hfill
    \begin{subfigure}{0.45\textwidth}
      \centering
      \includegraphics[width=\textwidth]{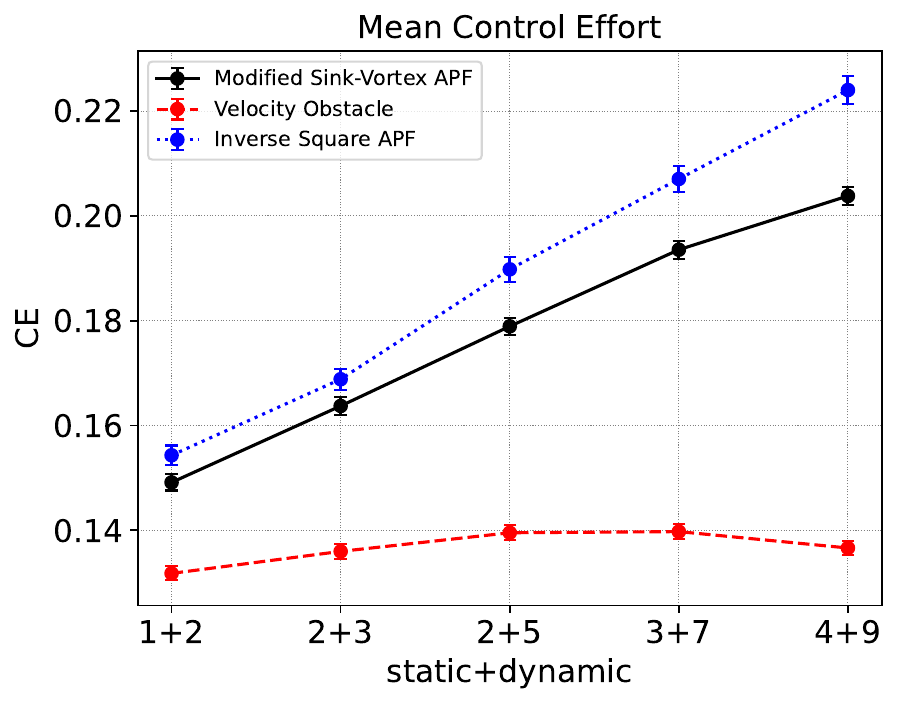}      
      \caption{Mean Cross Track Error (MCTE) computed using \eqref{eq:crosstrack_metric_equation}}
      \label{fig:statistics_cross_track_error}
    \end{subfigure}
    
    \caption{Statistical comparison of the modified sink-vortex APF with the traditional inverse square APF suggested by \cite{khatib1986real}. The bars for each data point represent the computed $95\%$ confidence interval from 5000 data points}. 
    \label{fig:statistical_comparison}
\end{figure*}

Simulations were performed over Amazon Web Services (AWS) cluster using the open source ray architecture \citep{moritz2018ray}. The cluster is configured with a head node of type m5.metal (96 virtual CPU cores and 384 GB memory) and with 3 workers of type m5.metal. The 384 core cluster allowed the 75000 simulations to be completed in less than three hours. The results of the statistical analysis are shown in \autoref{fig:statistical_comparison}.

It can be seen from \autoref{fig:statistics_success_rate} that the modified sink-vortex APF performs better than the traditional inverse square APF and the velocity obstacle method. The modified sink-vortex APF maintains over 99\% success rate in less congested environments. Even in the congested environment with four static and nine dynamic obstacles the method achieves a 97\% success rate. The confidence intervals denoted in the plot show that the statistical variation is small. It can be seen that both the traditional APF method and the velocity obstacle method see a decrease in success rate as the environment becomes more congested. It is important to note that while the modified sink-vortex APF method would attempt to follow COLREG rules by construction, the traditional APF and velocity obstacle method need not have followed COLREGS during these encounters. 

In the more congested environments (Environment 3, 4 and 5) the success rate of the velocity obstacle method is found to decrease significantly. It is important to note that the velocity obstacle method provides mathematical guarantees of collision avoidance only when the target ship is assumed to keep the same course and velocity. The velocity obstacle method is based on kinematics and does not take into account the dynamics of the interacting vessels. As real vessels cannot change direction or speed instantaneously, the mathematical guarantees do not hold when considering dynamics of the steering gear and the dynamics of the vessel. Similarly when both agents are taking actions to avoid collisions it cannot always be guaranteed that a collision will be avoided. 


\autoref{fig: statistics_computation} shows the computation cost for each of the methods. It is seen that both APFs have a significantly lower computation cost as compared to the velocity obstacle method. The velocity obstacle computation times increases by more than 100\% as the environment gets more congested. However, it is interesting to see that in more congested environments APFs do not incur a significant increase in computational cost. In more congested environments, each vessel encounters more number of vessels within their safe radius $R_{safe} = 15L$. In the velocity obstacle method, this causes an increase in the number of constraints on the selection of acceptable velocity vector leading to a greater computational cost. However, APF method relies only on summing up analytically evaluated gradients corresponding to each obstacle and hence the computation cost does not increase as significantly as observed in velocity obstacle method. 

It is also seen that computation cost per call of the modified sink-vortex APF is almost 25-30\% higher than the corresponding cost per call to the traditional APF. This increase is due to the computation of relative velocity in the modified sink-vortex APF method that is not required in the traditional inverse square APF method. However, as the computation time for all methods is only a few micro seconds, it does not have a significant effect on practical implementation.


\autoref{fig:statistics_averaged_time_to_reach_goal} shows that the mean time to reach the goal across the different environments for both APFs. It can be seen that the time to reach the goal increases in more congested enviroments for all three methods. The velocity obstacle method has a sharp increase in the mean time to reach goal as the number of obstacles increase. On the contrary the APF methods take significantly lower time than the velocity obstacle method and only show a mild increase in time to reach goal. 

For the traditional inverse square APF, when the goal is farther away than the obstacle, the repulsive gradients are dominant only when the vessel gets quite close to the obstacle. This causes the vessel's desired heading to change sharply to avoid collision and it takes considerable time for the vessel to get back towards the goal. On the other hand in the modified sink-vortex APF, heading change is initiated almost as soon as the obstacle is within the safe radius and causes the changes in heading to be smoother. This also allows the vessel to reach the goal in a lower time even in congested environments. As changes in heading are directly correlated with the applied rudder, it can be seen from \autoref{fig:statistics_control_effort} that the control effort due to the modified sink-vortex APF is significantly lower than the traditional APF. 

It is observed that the control effort associated with the velocity obstacle method is significantly lower than the APF methods. The velocity obstacle method aims to select a velocity that avoids collision with the other vessel and maintains this velocity until it steers clear of the other vessel. Maintaining a velocity only requires the application of rudder to achieve the desired heading and no further rudder is required after achieving the desired heading. Thus the control effort for the velocity obstacle method is consistently lower than the APF based methods.


\autoref{fig:statistics_cross_track_error} shows that the mean cross track error for both APFs increase in more congested environments. However, the mean cross track error due to modified sink-vortex APF is marginally higher in congested environments than the corresponding values observed from traditional inverse square APF. The velocity obstacle method leads to significantly larger mean cross track error as the deviation from the original path during an encounter with another vessel is significantly more.

\subsection{Comparison on a specific scenario}

\begin{figure*}
    \centering
    \begin{subfigure}{0.9\textwidth}
      \centering
      \includegraphics[width=0.5\textwidth]{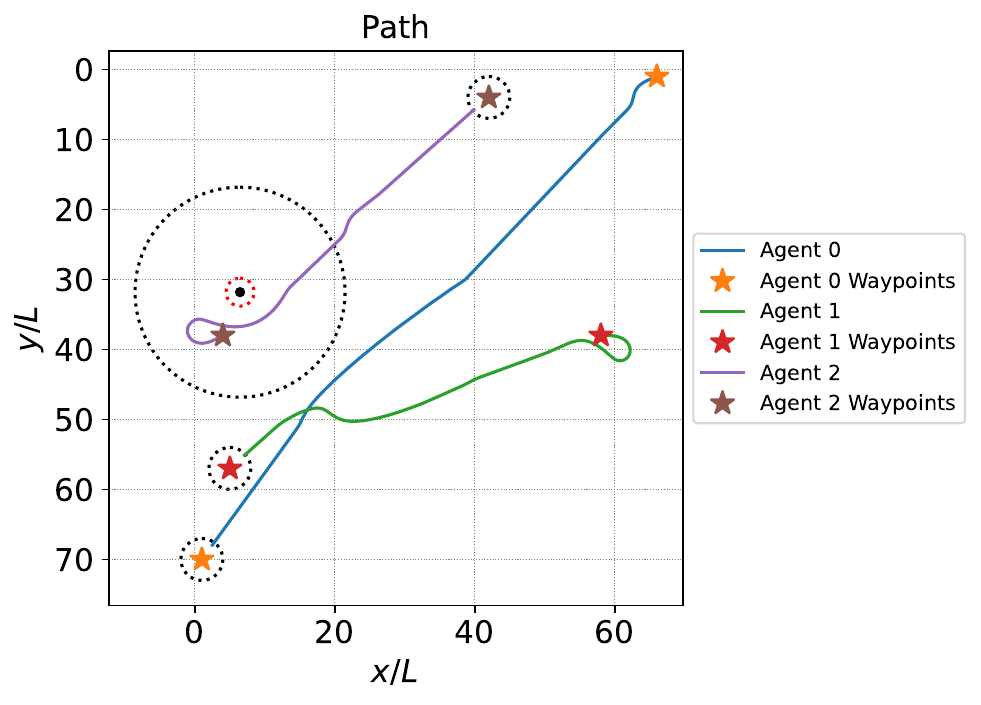}   
      \caption{Modified Sink-Vortex APF}
      \label{fig:env_01_modified_vortex_result}
    \end{subfigure}
    \begin{subfigure}{0.45\textwidth}
      \centering
      \includegraphics[width=\textwidth]{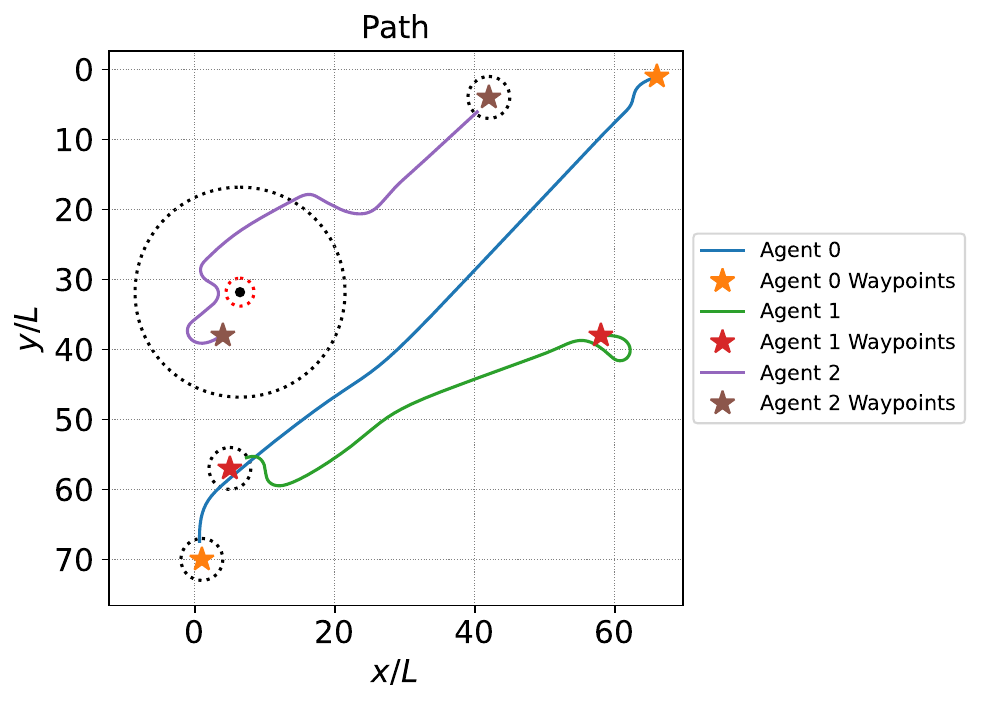}   
      \caption{Traditional Inverse Square APF}
      \label{fig:env_01_inverse_square_result}
    \end{subfigure}
    \begin{subfigure}{0.45\textwidth}
      \centering
      \includegraphics[width=\textwidth]{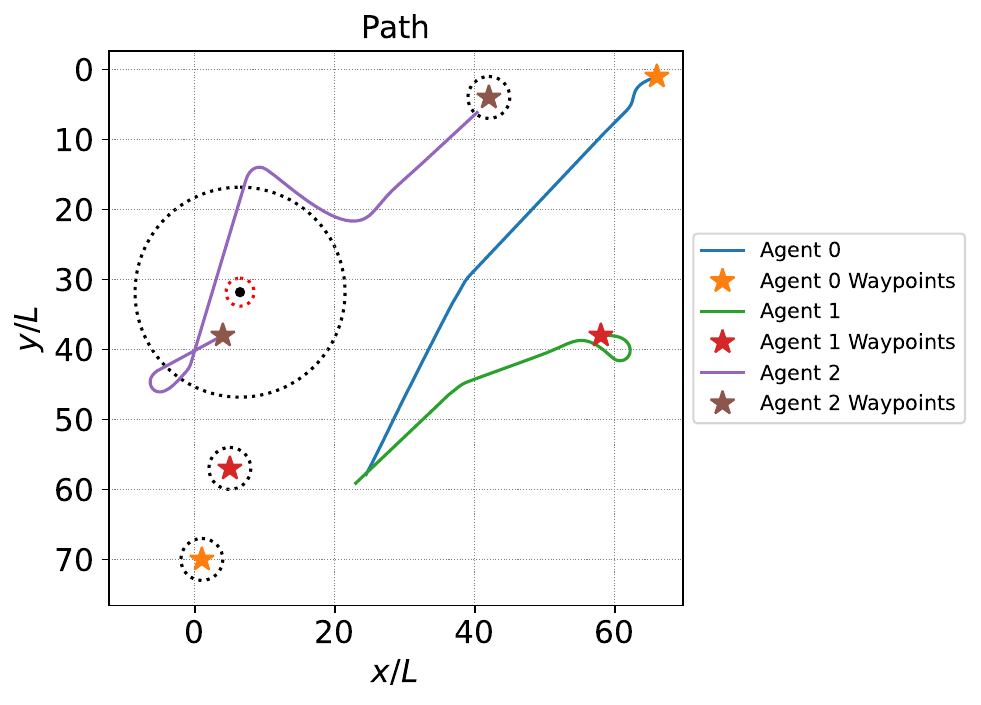}
      \caption{Velocity Obstacle}
      \label{fig:env_01_velocity_obstacle_result}
    \end{subfigure}
    \caption{Sample case from statistical analysis demonstrating collision avoidance demonstrated using the modified sink-vortex APF, the traditional APF architecture and the velocity obstacle method in Environment 1 with own ship (Agent 0), one static and two dynamic obstacles (Agent 1 and 2)}
    \label{fig:env_01_other_result}
\end{figure*}

In order to understand the advantages and limitations of all the methods, a single sample from the Environment 1 of the statistical analysis is simulated using all three methods. The paths taken by all vessels in the three methods are shown in \autoref{fig:env_01_other_result}. Note that each vessel is assumed to have the same length and the Froude numbers of each of the vessel in this scenario are:

\begin{enumerate}
    \item Ship 1 (Agent 0): $Fn = 0.2600$
    \item Ship 2 (Agent 1): $Fn = 0.1903$
    \item Ship 3 (Agent 2): $Fn = 0.2392$
\end{enumerate}

It can be seen from \autoref{fig:env_01_modified_vortex_result} that with the modified sink-vortex APF method, the first and second vessel have a crossing encounter and the third vessel avoids the static obstacle to reach its goal. Note that the second vessel (Agent 1) follows COLREG rules to gives way to the first vessel (Agent 0) during the encounter. Similarly, the third vessel (Agent 2) avoids the static obstacle through a starboard turn to reach the goal.

\autoref{fig:env_01_inverse_square_result} shows the results from the traditional APF method applied to the same scenario. It is seen that the first vessel and second vessel are repelled from each other during the later part of their voyage and maintain similar heading for a while. The first ship (Agent 0), being faster than the second vessel (Agent 1), crosses ahead of the second ship and after the crossing, the second ship takes a starboard turn to reach its goal. The third ship (Agent 2) avoids the static obstacle by taking a port turn.

\begin{figure}
    \centering
    \includegraphics[width=\linewidth]{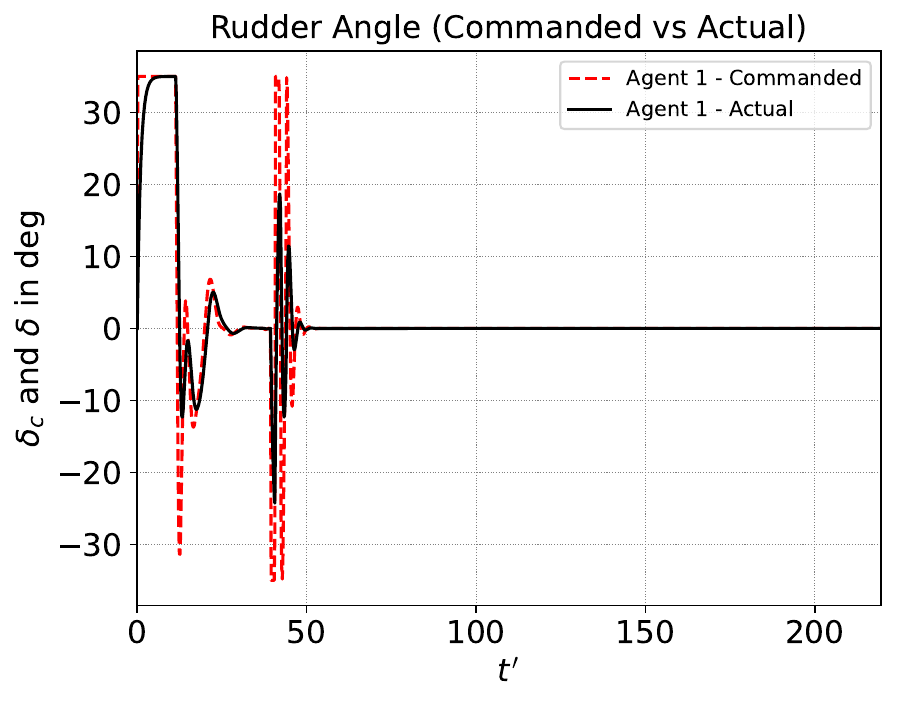}
    \caption{Rudder angle corresponding to \autoref{fig:env_01_velocity_obstacle_result}}
    \label{fig:env_01_velocity_obstacle_rudder}
\end{figure}

\autoref{fig:env_01_velocity_obstacle_result} shows the results from the velocity obstacle method. It is seen that the first vessel and the second vessel do change course as they approach each other. However, when they get quite close, it is observed that the desired heading for the second ship (Agent 1) fluctuates significantly leading to the commanded rudder to flip multiple times between $35^{\circ}$ and $-35^{\circ}$ as shown in \autoref{fig:env_01_velocity_obstacle_rudder}. However, the latency of the steering gear and the inertia of the vessel do not allow fast enough response leading to an eventual collision between the vessels. This failure to change course quickly leads to the vessels coming within the collision threshold of $2L$ distance of each other. It is also seen that the third vessel (Agent 2) takes a convoluted path to reach its destination.

This example demonstrates the ability of the proposed APF method to perform significantly better than the traditional APF method and the velocity obstacle method in terms of following COLREG rules and achieving a more natural collision free path to the goal. It is also seen that the deviation from the original course in minimal for the proposed APF.  
\rev{
\subsection{Limitations}

While the modified sink-vortex APF demonstrates a significantly better performance than the traditional APF and the velocity obstacle method, it does not achieve a 100\% success in all encounters. It is important to observe the failed simulations to understand the limitations of the approach. Two main mechanisms of failure were observed from the failed runs. 

\begin{figure}
    \centering    
    \includegraphics[width=\linewidth]{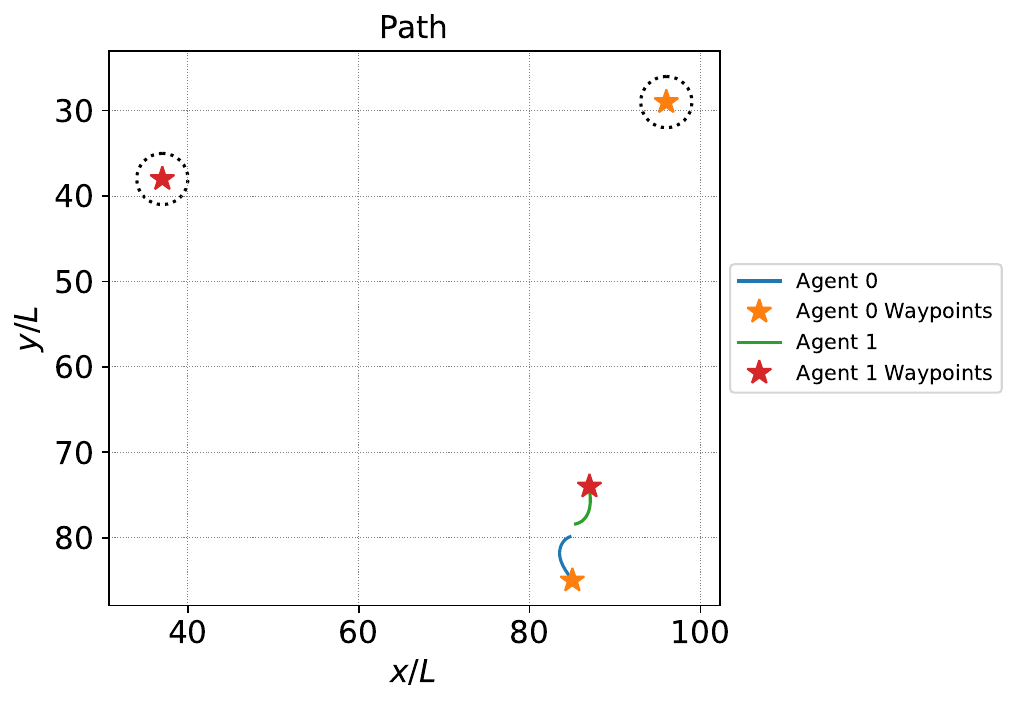}
      \caption{Extracted case depicting only colliding agents from a sample case taken from statistical simulation in environment 5}
    \label{fig:limitation_01}    
\end{figure}

\autoref{fig:limitation_01} shows a sample example taken from one of the statistical simulations of environment 5 where two ships are initialized with about $10L$ distance between them. It is seen that this separation is not sufficient enough to avoid collision. Even though both vessels aim to take starboard turns, they are unable to avoid a collision with each other due to their initial orientations. It is important to note that in reality ships will not suddenly appear in the vicinity of other ships with non-zero velocity and such scenarios are impractical. 

The second failure mechanism is observed when during an encounter between two ships, one of the vessels is very close to its goal waypoint. In this scenario, the attractive gradients towards the goal overshadow the repulsive gradients and lead the vessels to collide. Two such examples observed from the statistical analysis are shown in \autoref{fig:limitation_02}. 

\begin{figure*}
    \centering
    \begin{subfigure}{0.45\textwidth}
        \centering
        \includegraphics[width=\textwidth]{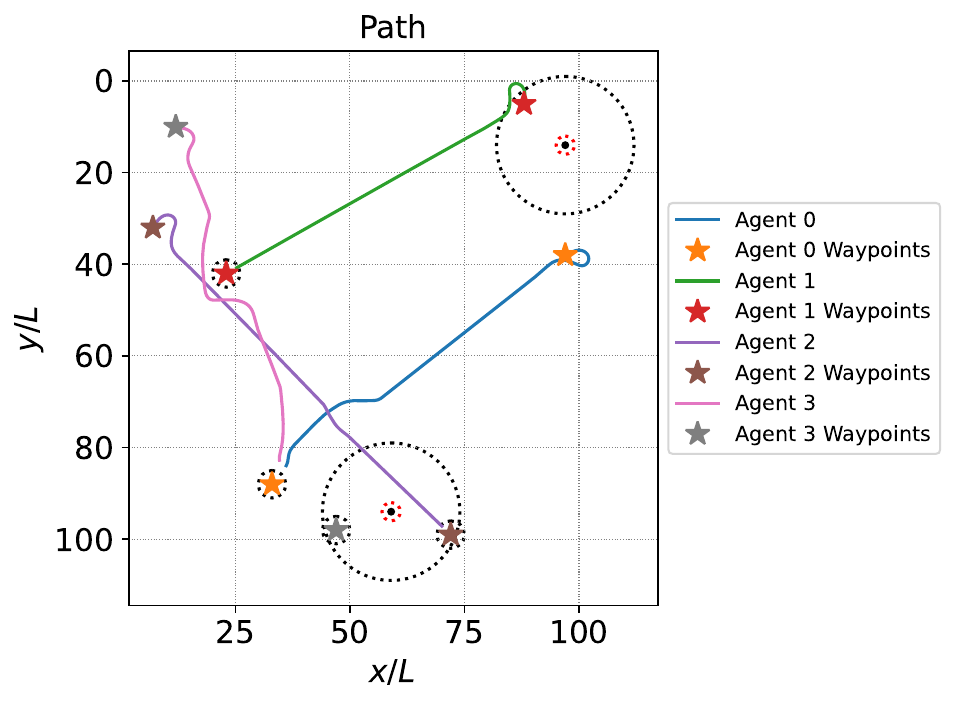}
        \caption{Environment 2 - Agent 0 and 3 collide}
    \end{subfigure}
    \begin{subfigure}{0.45\textwidth}
        \centering
        \includegraphics[width=\textwidth]{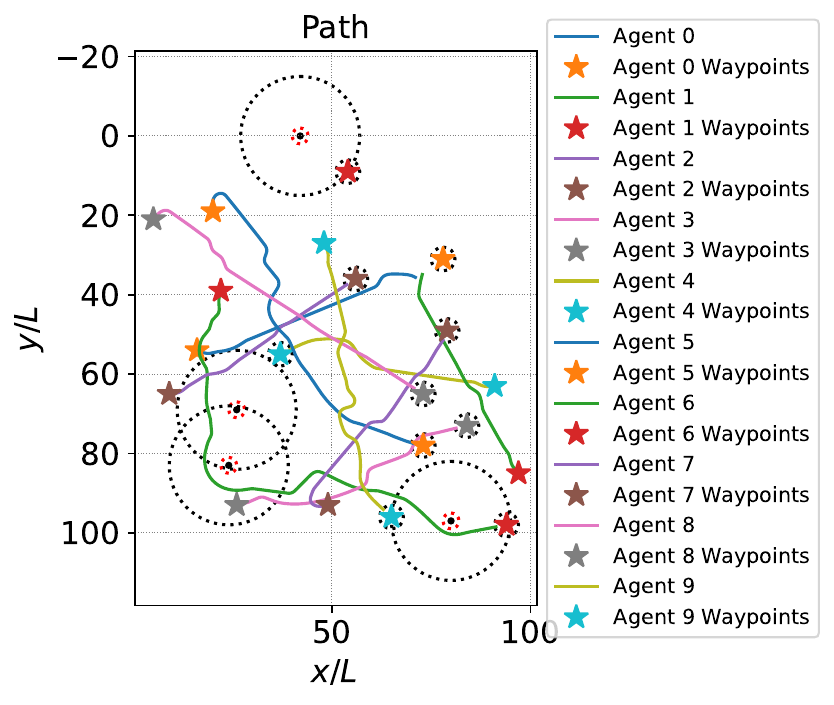}
        \caption{Environment 5 - Agent 0 and 6 collide}
    \end{subfigure}
    \caption{Failure mechanism when one of the vessel is too close to its goal}
    \label{fig:limitation_02}
\end{figure*}

While it is seen that increasing the safe radius leads to avoidance of collision in these specific examples, further investigations are needed to ascertain that this approach can lead to a statistically significant increase in success rate in congested environments.
}

\section{Conclusion and Future Work}
\label{sec:conclusion_futurework}

\rev{
This study proposes a novel artificial potential field (APF) based local planner to perform obstacle and collision avoidance in the maritime environment. Harmonic functions-based potential fields are proposed and further modified to achieve COLREGS compliance during multi-ship encounters. It is shown that the proposed APF is better suited for the maritime environments where the speed of the vessel cannot be changed suddenly and heading control is the primary approach to avoid collisions. It is also demonstrated that the method works well even in scenarios imposing conflicting COLREG responsibilities on the vessel. Further, the ability of the approach to maintain collision free navigation in narrow channels is also demonstrated. Finally, a statistical analysis is performed to test the proposed approach in a variety of environments and it is seen that the method has a significantly higher success rate than both the traditional APF method and the velocity obstacle method. Further comparisons with velocity obstacle method show that the proposed APF is computationally inexpensive and results in significantly lower cross track error. The proposed approach also beats the velocity obstacle method in terms of the time taken to reach the goal by a 10-30\% margin. Thus, the proposed approach would serve as a good pivoting point in developing algorithms for fully autonomous ships that aim to follow COLREG rules by construction.

In the future, experimental analysis of the proposed APF will undertaken on free running scaled models of ships to validate the approach in field. This will also allow the relaxation of the assumption of perfect knowledge of the state of the vessel assumed in the current study and understand its effect on the reactive guidance. In the statistical analysis it is seen that about 3\% of the encounters in congested environments still result in a collision. Although a greater safe radius has been seen to reduce this percentage, studies will be undertaken to suggest further modifications to the approach to reduce these collisions while keeping the safe radius same. 
}

\section*{Acknowledgement}

This work was partially funded by the Science and Engineering Research Board (SERB) India - SERB Grant Number CRG/2020/003093 and the New Faculty Seed Grant of IIT Madras. 
This work is also supported through the \href{https://ioe.iitm.ac.in/project/marine-autonomous-systems/}{Center of Excellence for Marine Autonomous Systems (CMAS), IIT Madras} that is now a part of the Center for Maritime Experiments to Maritime Experience (ME2ME) setup under the Institute of Eminence Scheme of Government of India.

\printcredits

\bibliographystyle{cas-model2-names}

\bibliography{references}

\end{document}